%% file: main.tex
\documentclass{article}

\usepackage[table,xcdraw]{xcolor}  % before arxiv.sty; `table' loads colortbl
\usepackage{styles/arxiv}

\usepackage[T1]{fontenc}
\usepackage{microtype}
\usepackage{enumitem}              % itemize options in the appendix
\usepackage{fontawesome5}          % envelope / GitHub icons in the author block
\usepackage{math_commands}
\usepackage{tikz}                  % fig_cases_v3_tikz.tex
\usepackage[most]{tcolorbox}       % `most' = skins (enhanced), breakable, listings, ...

\usepackage{makecell}
\usepackage{siunitx}
\usepackage{pifont}
\usepackage{caption}
\definecolor{sentrybg}{HTML}{EAF1FB}   % light blue row shading for Sentry rows
\definecolor{gainc}{HTML}{1E7B34}      % gains
\definecolor{dropc}{HTML}{B3261E}      % drops
\definecolor{subtle}{gray}{0.42}       % metric labels, helper rows

\newcommand{\pmstd}[1]{{\scriptsize$\,\pm\,$#1}}           % small std after a mean
\newcommand{\second}[1]{\underline{#1}}                     % second best
\newcommand{\metric}[1]{{\footnotesize\color{subtle}#1}}    % metric line under benchmark name
\newcommand{\gain}[1]{{\footnotesize\color{gainc}$+$#1\%}}
\newcommand{\drop}[1]{{\footnotesize\color{dropc}$-$#1\%}}
\newcommand{\dropabs}[1]{{\footnotesize\color{dropc}$-$#1}}
\newcommand{\grouprow}[2]{\multicolumn{#1}{l}{\footnotesize\itshape\color{subtle}#2}}
\newcommand{\cmark}{\ding{51}}
\newcommand{\xmark}{{\color{subtle}\ding{55}}}

\input{macros.tex}

\title{Sentry: Learning to Recover from LLM Agent Failures at Test Time}

\author{\textbf{Changxiu Ji}
\textsuperscript{}$^*$
\quad
\textbf{Amy Lu}
\textsuperscript{}$^*$
\quad
\textbf{Qizheng Zhang}
\quad
\textbf{Kunle Olukotun}\\[5pt]
Department of Computer Science, Stanford University \quad $^*$ equal contribution \\[3pt]
\faEnvelope \enspace \texttt{leo0610@stanford.edu, qizhengz@stanford.edu} \quad
\faGithubSquare \enspace \texttt{\href{https://github.com/nuglifeleoji/Sentry}{Sentry}}
}

\begin{document}

\maketitle

\input{sections/abstract_final}
\input{sections/introduction_final}
\input{sections/background_final}
\input{sections/method_final}
\input{sections/setup_final}
\input{sections/results_final}
\input{sections/discussion_final}

\bibliographystyle{plain}
\bibliography{references}

%%%%%%%%%%%%%%%%%%%%%%%%%%%%%%%%%%%%%%%%%%%%%%%%%%%%%%%%%%%%
\clearpage

\appendix
\input{sections/appendix_final}

%%%%%%%%%%%%%%%%%%%%%%%%%%%%%%%%%%%%%%%%%%%%%%%%%%%%%%%%%%%%

\end{document}

%% file: macros.tex
\newcounter{packednmbr}

\newenvironment{packeditemize}{\begin{list}{$\bullet$}{\setlength{\itemsep}{0.5pt}\addtolength{\labelwidth}{-4pt}\setlength{\leftmargin}{2ex}\setlength{\listparindent}{\parindent}\setlength{\parsep}{1pt}\setlength{\topsep}{2pt}}}{\end{list}}

\definecolor{refinegreen}{RGB}{0, 128, 75}
\definecolor{scoregreen}{RGB}{34, 139, 34}

\definecolor{darkgray}{RGB}{70,70,70}
\definecolor{lightgray}{RGB}{240,240,240}

\tcbset{
    boxstyle/.style={
        enhanced,
        sharp corners,
        colback=gray!20,
        colframe=gray!60,
        boxrule=1pt,
        left=10pt,
        right=10pt,
        top=5pt,
        bottom=5pt
    },
    headerstyle/.style={
        enhanced,
        sharp corners,
        colback=gray!60,
        coltext=white,
        boxrule=0pt,
        left=10pt,
        right=10pt,
        top=5pt,
        bottom=5pt,
        fontupper=\bfseries
    },
    verbatimstyle/.style={
        enhanced,
        sharp corners,
        colback=gray!10,
        colframe=gray!40,
        boxrule=1pt,
        left=10pt,
        right=10pt,
        top=5pt,
        bottom=5pt,
        listing only,
        fontupper=\ttfamily\small
    }
}

\newtcolorbox{examplebox}[1][]{
    colback=lightgray!10,
    colframe=black,
    boxrule=0.75pt,
    title=#1,
    fonttitle=\bfseries,
    left=3pt,
    right=3pt,
    top=2pt,
    bottom=2pt,
    breakable,  % Allow the box to span pages
}

\newtcolorbox{exampleboxcode}[1][]{
    colback=lightgray!10,
    colframe=blue,
    boxrule=0.75pt,
    title=#1,
    fonttitle=\ttfamily,
    left=3pt,
    right=3pt,
    top=2pt,
    bottom=2pt,
}

\definecolor{codegreen}{rgb}{0,0.6,0}
\definecolor{codegray}{rgb}{0.5,0.5,0.5}
\definecolor{codepurple}{rgb}{0.58,0,0.82}
\definecolor{backcolour}{rgb}{0.97,0.97,0.97}

\lstdefinestyle{codestyle}{
    commentstyle=\color{codegreen},
    keywordstyle=\color{magenta},
    numberstyle=\tiny\color{codegray},
    stringstyle=\color{codepurple},
    basicstyle=\ttfamily\scriptsize,
    breakatwhitespace=false,         
    breaklines=true,                 
    captionpos=b,                    
    keepspaces=true,                 
    numbers=left,                    
    numbersep=5pt,                  
    showspaces=false,                
    showstringspaces=false,
    showtabs=false,                  
    tabsize=2,
    xleftmargin=10pt,                % Left margin
    xrightmargin=10pt                % Right margin
}

%% file: sections/abstract_final.tex
\begin{abstract}
LLM agents often fail mid-task due to invalid tool calls, repeated actions, or poorly grounded reasoning, and learning from these failures is a path to reliability.
We find that \emph{how} failure knowledge reaches the agent matters as much as \emph{what} it contains.
Failure lessons are conditional: kept in the agent's context, they misfire when their failure is absent, and removing them from an evolving playbook improves performance.
Runtime interventions, in contrast, act only when a failure occurs but do not learn from their repairs.
We argue that failure knowledge is conditional knowledge and should be conditionally exposed, and instantiate this principle in Sentry, a failure-management layer that runs alongside the agent.
When Sentry detects a failure, it retrieves matching lessons from an external playbook to guide recovery, verifies without access to task rewards whether the agent recovered, and stores a new lesson only if it did; the full playbook never enters the agent's context.
Across multiple agentic benchmarks, Sentry outperforms the strongest runtime-intervention baseline on every benchmark, by 37\% on average, and the strongest context-evolution baseline by 39\% on the two benchmarks where both are evaluated; combining Sentry with context evolution yields further gains.
Learned lessons transfer to held-out tasks, and controlled experiments show that exposing the full playbook to the agent lowers performance even when relevant lessons remain available on demand.
\end{abstract}

%% file: sections/introduction_final.tex
\section{Introduction}
\label{sec:intro}

Modern large language model (LLM) agents increasingly tackle interactive and multi-step tasks involving reasoning, tool use, and external environments~\citep{yao2023react,schick2023toolformer,yao2024taubench,merrill2026terminalbenchbenchmarkingagentshard}.
Success often hinges less on getting every step right than on recovering when one goes wrong: an invalid tool call~\citep{winston2025taxonomy}, a repeated or stalled action, a poorly grounded decision~\citep{zhu2025where}, or reasoning that does not match the action~\citep{cemri2025mast}.
As tasks grow longer, failure recovery becomes central to agent reliability.

Two lines of work let agents learn from failures, and each gets only half the problem right.
\textit{Context-evolution} methods distill past trajectories into reflections, workflows, or playbooks reused on future tasks~\citep{shinn2023reflexion,wang2025awm,ouyang2026reasoningbank,zhang2026ace}.
They learn across tasks but expose lessons too broadly: even retrieval-based memories~\citep{zhao2023expel,ouyang2026reasoningbank} select lessons once at task start, before any failure, and keep them in context for the whole trajectory.
\textit{Runtime-intervention} methods monitor the trajectory and inject guidance when a failure is detected~\citep{koohestani2025agentguard,zhang2026agentforesight,nanda2026wink,mulian2026agentfixer}.
They act selectively but do not learn: what resolved one failure does not inform the next, and a fixed rule that rescues one task can derail another (Figure~\ref{fig:failure_knowledge_cases}b).

\textbf{We find that \emph{how} failure knowledge reaches the agent matters as much as \emph{what} it contains.}
Failure lessons are conditional:``if repeated searches return nothing relevant, broaden the query" helps only when that failure is occurring.
Kept in context, such lessons force the agent to judge at every step whether each applies, and it often judges wrong: in Figure~\ref{fig:failure_knowledge_cases}a, an agent carrying recovery rules learned by ACE~\citep{zhang2026ace} finds a matching product but keeps comparing alternatives and never buys it.
The effect is systematic.
Removing failure lessons from ACE's playbook \emph{improves} held-out performance, from 0.335 to 0.391 reward on WebShop~\citep{yao2022webshop} and from 0.562 to 0.643 completion on Mind2Web Replay~\citep{deng2023mind2web}.
Conversely, adding a fixed playbook to persistent context lowers WebShop reward from 0.364 to 0.296, even when the same lessons remain available on demand, mirroring capability erosion in self-evolving agents~\citep{yu2026selfevolvingforget}.
Intervention shows the same pattern: intervening without checking whether it is needed drops WebShop reward below the unassisted agent's (0.136 vs.\ 0.170).
These observations suggest a simple principle: \textbf{failure knowledge is conditional knowledge and should be conditionally exposed.}
It should be \emph{learned} across tasks, as in context evolution, but \emph{surfaced} only on a matching failure, as in runtime intervention.

We instantiate this principle in \textbf{\textit{Sentry}}, a failure-management layer that runs alongside the task agent and learns from the recoveries it verifies.
Sentry watches the agent's recent steps and returns non-executable actions for a schema-conforming retry.
For behavioral failures, such as loops or reasoning not grounded in observations, Sentry diagnoses the failure, retrieves at most $k$ matching lessons from an external playbook, and uses them to guide the agent's next step.
It then checks whether the agent recovers, without access to task rewards, and writes a new lesson only if it does.
The full playbook never enters the agent's context: the agent sees a handful of lessons, only while failing, and only for that failure.
Because the problem stems from how knowledge enters the context rather than from a particular model's weaknesses, we expect it to matter more, not less, as memories grow and agents run longer.
 
Across WebShop, AppWorld~\citep{trivedi2024appworld}, SWE-bench Lite~\citep{jimenez2024swebench}, and Mind2Web Replay, Sentry outperforms the strongest runtime-intervention baseline on every benchmark, by 37\% on average (up to 77\%), and ACE, the strongest context-evolution baseline, by 39\%.
The gains are not just action repair: without the schema-conforming retry, Sentry still beats all runtime-intervention baselines on both ablated benchmarks.
Sentry also complements context evolution: ACE+Sentry performs best wherever evaluated, suggesting that task- and failure-level knowledge can evolve together.
Our contributions are:
\begin{packeditemize}
    \item \textbf{A finding about failure knowledge.} We show that failure-specific knowledge is conditional, and that exposing it unconditionally hurts: always-on failure lessons and always-on intervention both lower agent performance in controlled comparisons (\S\ref{sec:background}, \S\ref{sec:ablations}).
 
    \item \textbf{Sentry, a method built on this finding.} Sentry keeps an evolving recovery playbook outside the agent's context, retrieves from it only when a matching failure is diagnosed mid-trajectory, and admits a lesson only after verifying, without reward access, that the recovery succeeded (\S\ref{sec:method}).
 
    \item \textbf{Evidence across four benchmarks.} Beyond the main comparisons, held-out evaluation shows that learned recovery lessons transfer to unseen tasks; paired forks from 200 intervention points show that Sentry's guidance causally improves local recovery (33.5\% $\rightarrow$ 60.5\%); and human annotation agrees with Sentry's intervention and recovery judgments in 86.0\% and 81.8\% of cases (\S\ref{sec:results}).
\end{packeditemize}

%% file: sections/background_final.tex
\section{Background and Motivation}
\label{sec:background}

Methods that let LLM agents learn from execution failures differ along two axes: \emph{whether} failure knowledge accumulates across tasks, and \emph{what it is conditioned on} when it reaches the agent.
We examine both families along these axes; Figure~\ref{fig:failure_knowledge_cases} shows how each goes wrong on WebShop with Qwen3.5-9B (full trajectories in Appendix~\ref{app:failure_knowledge_cases}; further related work in Appendix~\ref{app:related_work}).

\paragraph{Context evolution conditions on the task, but failures happen mid-trajectory.}
Reflexion stores natural-language reflections~\citep{shinn2023reflexion}, AWM induces reusable workflows~\citep{wang2025awm}, and ACE maintains an evolving playbook of advice~\citep{zhang2026ace}; ExpeL and ReasoningBank retrieve a subset of lessons for each new task~\citep{zhao2023expel,ouyang2026reasoningbank}.
In all cases, lessons are selected before execution and stay in context throughout the task.
This is the wrong conditioning variable for failure knowledge: whether ``broaden the query after repeated empty searches'' applies depends on the agent's state at a particular step, which is unknown when lessons are selected.
Better task-level retrieval cannot fix this, and the agent is left to judge at every step which lessons apply.
In Figure~\ref{fig:failure_knowledge_cases}(a), three such rules keep the agent comparing alternatives after it has already found a matching product.

\paragraph{Runtime intervention conditions on the state, but cannot learn its own boundaries.}
AgentGuard warns the agent when fixed rules fire, e.g., on invalid actions or no-progress steps~\citep{koohestani2025agentguard}; AgentForesight raises early alarms from partial trajectories~\citep{zhang2026agentforesight}; Wink uses an LLM monitor to suggest course corrections~\citep{nanda2026wink}; and AgentFixer combines rule-based checks with LLM judges~\citep{mulian2026agentfixer}.
These methods condition on the right variable, the current trajectory, but their knowledge is fixed.
Each intervention is written from scratch or from a hand-set rule, so nothing tells the system when an intervention helped, when it hurt, or how to tell the two apart next time.
In Figure~\ref{fig:failure_knowledge_cases}(b), ``page unchanged'' signals a genuine stall in one task and valid option selection in another; AgentGuard's rule cannot learn the difference.

\paragraph{Always-on failure knowledge costs more than it helps.}
To isolate the cost of exposure, we remove failure-mode lessons from ACE's playbook under the held-out protocol of Table~\ref{tab:self_evolving_baselines} (memory initialized on 50 tasks, evaluated on 50 disjoint tasks; \S\ref{protocol}).
Pruning improves ACE on both WebShop and Mind2Web Replay, our interactive adaptation of Mind2Web~\citep{deng2023mind2web} (\S\ref{sec:benchmarks_metrics}).
On average, then, the harm these lessons cause when their failure is absent outweighs the help they give when it occurs.
Pruning is a blunt fix, however: it discards that help along with the harm.
The alternative is to keep the lessons but change \emph{when} the agent sees them: with the same playbook, failure-triggered exposure outperforms persistent exposure (\S\ref{sec:ablations}).

\begin{figure}[t]
    \centering
    \includegraphics[width=0.8\linewidth]{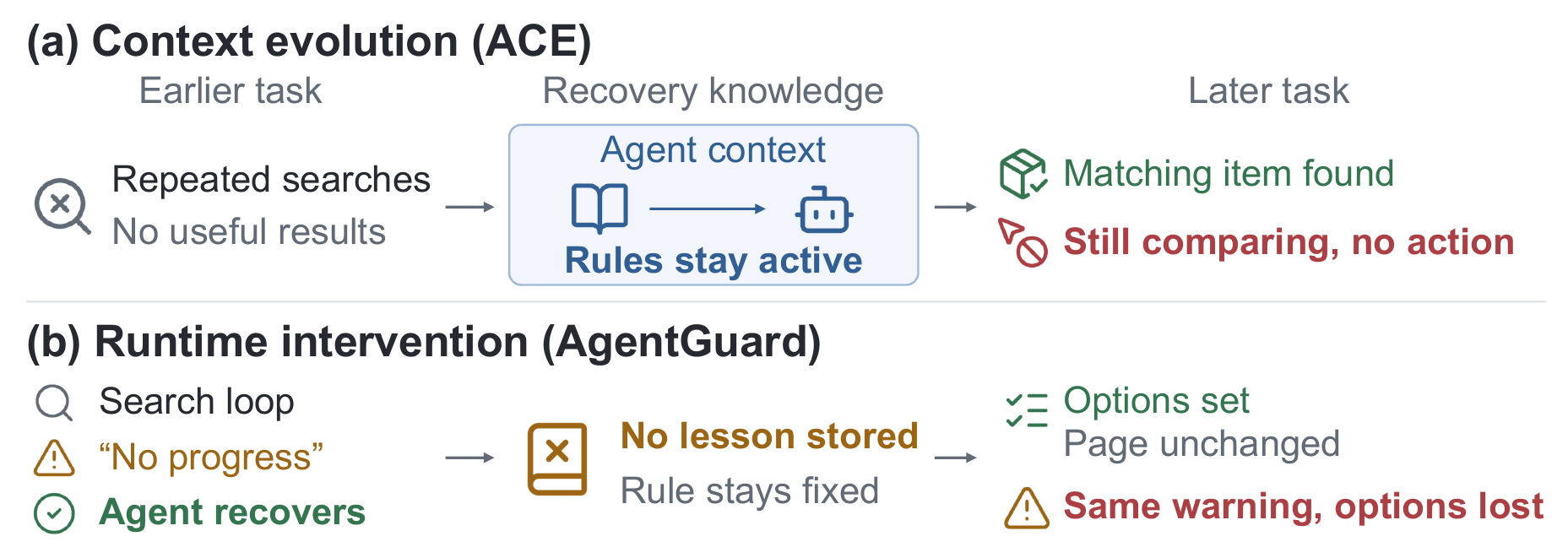}
    \caption{\textbf{Two ways failure knowledge goes wrong} (Qwen3.5-9B on WebShop).
    (a) \emph{Always exposed:} recovery rules learned by ACE stay in the agent's
    context; after finding a matching item, the agent keeps comparing
    alternatives and takes no action.
    (b) \emph{Never learned:} AgentGuard's fixed no-progress warning resolves a
    search loop, but no lesson is stored; in a later task, the same warning
    fires after valid option selections, and the agent leaves the page and
    loses them.}
    \vspace{-7pt}
    \label{fig:failure_knowledge_cases}
\end{figure}

\paragraph{Takeaway.}
Failure knowledge should be \emph{learned} across tasks, as in context evolution; \emph{surfaced} only when a failure is diagnosed mid-trajectory, as in runtime intervention; and \emph{admitted} only once its recovery is verified, since a lesson learned from a failed repair would misfire on future tasks too.
Sentry (\S\ref{sec:method}) is built around these three properties.

%% file: sections/method_final.tex
\section{Sentry: A Framework for Test-Time LLM Agent Failure Recovery}
\label{sec:method}

\begin{figure}[t]
    \centering
    \includegraphics[width=0.95\linewidth]{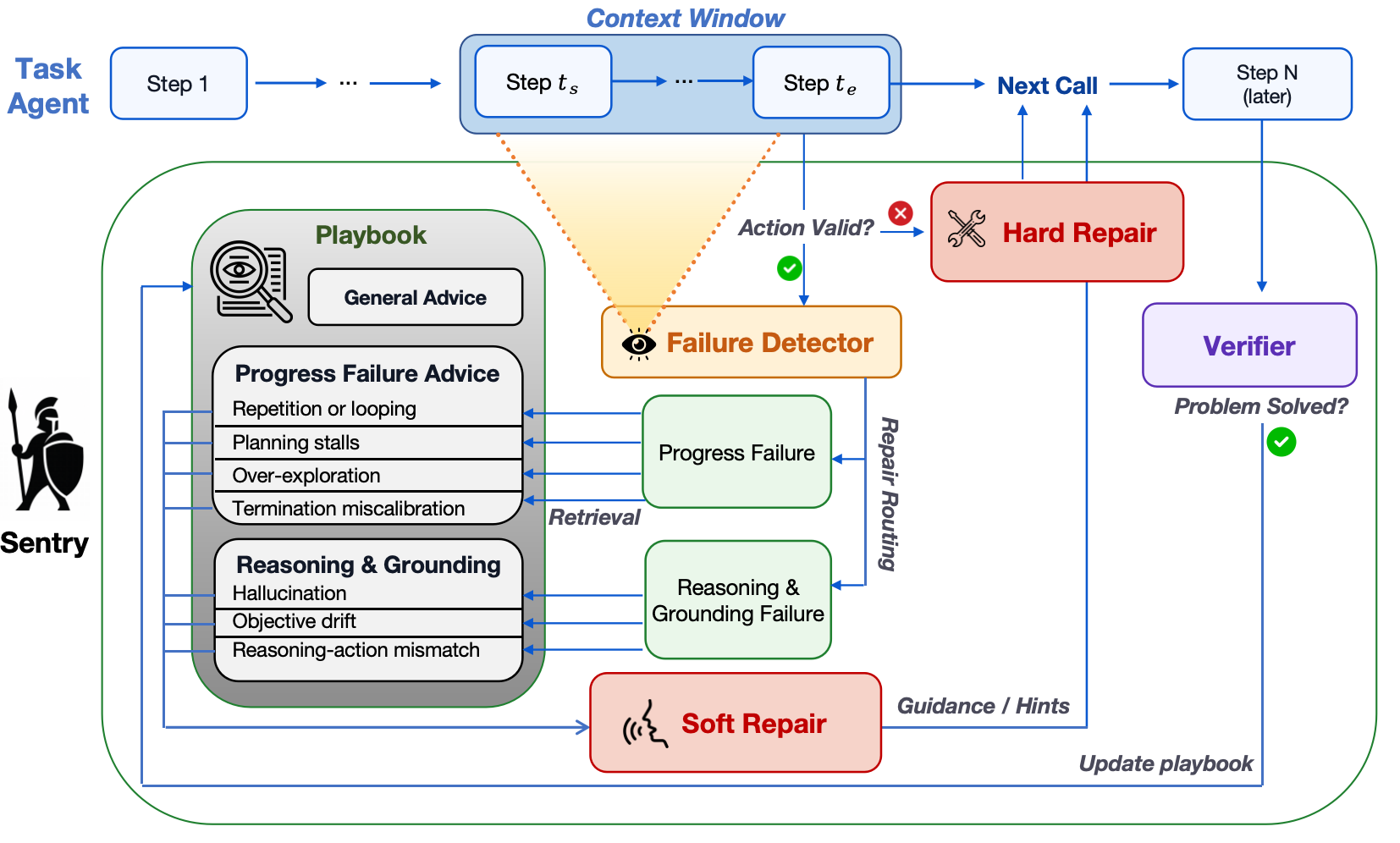}
    \caption{\textbf{Sentry's framework.} Sentry routes invalid actions to hard repair and progress or reasoning--grounding failures to soft repair with retrieved playbook guidance. Hard repairs are checked immediately; soft repairs undergo delayed recovery verification. Only verified soft-repair recoveries are added to the external playbook for use in future interventions.}
    \vspace{-7pt}
    \label{fig:sentry}
\end{figure}

We introduce \textbf{\textit{Sentry}} (Figure~\ref{fig:sentry}), a failure-management layer that runs alongside the task agent and handles failure knowledge along the three properties of \S\ref{sec:background}: it \emph{surfaces} recovery lessons only when it diagnoses a failure (\S\ref{sec:detection}, \S\ref{sec:soft_repair}), \emph{admits} a lesson only after verifying the recovery, and \emph{learns} by accumulating verified lessons across tasks (\S\ref{sec:verification}).
The complete playbook is maintained outside the agent's persistent background context; we refer to this as \emph{isolation}.
Isolation does not hide recovery knowledge from the agent: the few entries retrieved for a detected failure still enter the conversation during that intervention.

During execution, Sentry monitors the agent's recent trajectory.
It routes invalid actions to \textit{hard repair}, which asks the task agent for a valid replacement action, and routes progress and reasoning--grounding failures to \textit{soft repair}, which combines the local diagnosis with relevant playbook entries to guide the agent's next step.
After a soft repair, Sentry verifies whether the triggering failure has been resolved and adds only verified recovery lessons to the playbook for later interventions.
Prompt templates are provided in Appendix~\ref{prompts}.

\subsection{Failure detection and routing}
\label{sec:detection}

At each step, Sentry decides whether to intervene and, if so, which repair route to take.
Given the task objective, it examines the current action and the agent's $W = 5$ most recent reasoning--action--observation cycles.

Sentry first checks whether the action is executable under the environment's required schema.
If it is not, Sentry invokes \textit{hard repair}: the task agent receives the invalid action, the relevant trajectory context, and the schema, and is instructed to output only a corrected action in the required format.

If the action is valid, Sentry checks the recent trajectory for two types of failures: \textit{Progress Failures}, such as repetition or stalled execution, and \textit{Reasoning \& Grounding Failures}, such as unsupported claims or reasoning--action mismatch.
For a detected failure, Sentry produces a diagnosis consisting of the failure type, fine-grained retrieval labels, and supporting local evidence, which then triggers \textit{soft repair}.
If no failure is detected, execution continues with no intervention and no playbook retrieval, so the agent sees no recovery knowledge while it is making progress.
Full failure and retrieval-label definitions are provided in Appendix~\ref{app:retrieval_labels}.

\subsection{Soft repair with playbook retrieval}
\label{sec:soft_repair}

Each playbook entry is a record of a verified recovery: its failure type, retrieval labels, trigger pattern, and repair principle.
Given a diagnosis, Sentry selects entries with the same failure type and at least one shared retrieval label, ranks them by label overlap, and retrieves at most $k = 5$.
If no entry shares a retrieval label, Sentry falls back to the most recent entries of the same failure type.
The procedure, including recency tie-breaking and label-diversity constraints, is provided in Appendix~\ref{app:playbook_retrieval}.
 
Sentry then builds a \textit{soft-repair} prompt from the retrieved lessons, the failure type, the supporting local evidence, and general repair instructions.
The prompt guides the task agent to address the failure and choose a productive next step while staying on the original task objective.
When the playbook is empty, soft repair relies on the general repair instructions alone; as verified rescues accumulate, retrieved lessons increasingly shape the guidance.
Because retrieval is bounded, the playbook can grow across tasks while each intervention exposes the agent to at most $k$ lessons.

\subsection{Recovery verification and playbook update}
\label{sec:verification}

Before any experience enters the playbook, Sentry verifies that the intervention resolved the detected failure.
For hard repair, verification is immediate: the corrected action must satisfy the required schema and be accepted by the environment.
For soft repair, Sentry examines the subsequent $H = 10$ reasoning--action--observation cycles and judges the failure resolved when the agent stops the diagnosed failure pattern and resumes task-relevant, observation-grounded progress.
This judgment concerns recovery from the triggering failure only; the verifier has no access to benchmark rewards or final task-success labels.
 
Only soft-repair episodes judged resolved contribute new playbook entries; unresolved episodes produce none.
For each resolved episode, Sentry summarizes the diagnosis, repair guidance, and observed recovery into a reusable record with the failure type, retrieval labels, trigger pattern, and repair principle.
Verification thus controls \emph{which} experiences enter the playbook, while failure-conditioned retrieval controls \emph{when} they are activated.
Self-evolution occurs only through playbook updates; the detection and retrieval procedures remain fixed.
Appendix~\ref{playbook} provides an example playbook produced by Sentry.

%% file: sections/setup_final.tex
\section{Experimental Setup}
\subsection{Benchmarks and Metrics}
\label{sec:benchmarks_metrics}

\paragraph{Benchmarks.}
We evaluate on WebShop and Mind2Web Replay (web agents), SWE-bench Lite
(software engineering), and AppWorld (stateful app-based tool use)
\citep{yao2022webshop,deng2023mind2web,jimenez2024swebench,trivedi2024appworld}.
Mind2Web Replay is our interactive adaptation of Multimodal-Mind2Web: after a
rejected action, the agent remains at the reference state and may retry
within its execution budget, so the benchmark measures feedback-driven recovery
along trajectories. All methods share the same candidate-based
interface and feedback; scores are not comparable to the official Mind2Web
step-wise metrics (Appendix~\ref{app:Benchmarks and Metrics}). The
runtime-intervention comparison uses 100 tasks on WebShop, Mind2Web Replay,
and AppWorld, and 50 on SWE-bench Lite.

\paragraph{Metrics.}
We report average reward on WebShop; mean normalized trajectory completion, the
fraction of reference steps completed in order, on Mind2Web Replay;
issue-resolution accuracy under the official harness on SWE-bench Lite; and test
pass rate, the percentage of checks passed per task, on AppWorld.
Secondary metrics and intervention counts are in
Appendices~\ref{app:Benchmarks and Metrics} and~\ref{app:full_main_results}.

\subsection{Baselines}

The \textbf{Base agent} runs the model without runtime intervention or
evolution. Runtime-intervention baselines are
\textbf{AgentGuard}~\citep{koohestani2025agentguard},
\textbf{AgentFixer}~\citep{mulian2026agentfixer},
\textbf{Wink}~\citep{nanda2026wink}, and
\textbf{AgentForesight}~\citep{zhang2026agentforesight}; context-evolution
baselines are \textbf{Reflexion}~\citep{shinn2023reflexion},
\textbf{ACE}~\citep{zhang2026ace}, \textbf{AWM}~\citep{wang2025awm}, and
\textbf{ReasoningBank}~\citep{ouyang2026reasoningbank}. We also evaluate
\textbf{ACE+Sentry}, which adds Sentry's failure manager to ACE.
Implementations are in Appendix~\ref{baseline}.

\subsection{Implementation Details and Evaluation Protocols}
\label{protocol}

By default, the task agent and all Sentry components use
Qwen3.5-9B~\citep{qwen2026qwen35}; we also evaluate
GPT-OSS-120B~\citep{openai2025gptoss}. Task sets are sampled once from the
executable tasks of each official split and held fixed across five runs with
randomized execution order; we report mean $\pm$ standard deviation over these
runs. Decoding settings, output schemas, and task-selection details are in
Appendices~\ref{prompts} and~\ref{app:implementation_details}.
 
\paragraph{Evaluation protocols.}
The runtime-intervention comparison and component ablations use an
\textit{online-from-scratch} protocol: Sentry starts with an empty playbook and
accumulates verified recoveries online. Context-evolution comparisons and the
playbook exposure, transfer, and retrieval analyses use a \textit{held-out}
protocol on WebShop and Mind2Web Replay: each method initializes its memory on
50 tasks and is evaluated on 50 disjoint tasks, with set membership fixed
across runs. During evaluation, context-evolution baselines keep updating after
each task and Sentry after each verified recovery; only held-out tasks are
scored. Retrieval details are in Appendix~\ref{app:playbook_retrieval}.

%% file: sections/results_final.tex
\section{Results}
\label{sec:results}

\subsection{Main results}
\label{sec:main_results}

\begin{table}[t]
\centering
\caption{\textbf{Comparison with runtime-intervention baselines.}
All methods use Qwen3.5-9B and the online-from-scratch protocol (\S\ref{protocol});
we report mean $\pm$ std over five runs.
\textbf{Bold}: best per column; \underline{underline}: second best.
The last row gives Sentry's relative gain over the strongest baseline in each column.}
\label{tab:main_results_all}
\small
\setlength{\tabcolsep}{6pt}
\renewcommand{\arraystretch}{1.12}
\begin{tabular}{lcccc}
\toprule
Method
& \makecell{WebShop\\\metric{Avg.\ reward $\uparrow$}}
& \makecell{AppWorld\\\metric{Test pass (\%) $\uparrow$}}
& \makecell{SWE-bench Lite\\\metric{Accuracy $\uparrow$}}
& \makecell{Mind2Web Replay\\\metric{Completion $\uparrow$}} \\
\midrule
Base agent
& 0.170\pmstd{0.016} & 25.59\pmstd{0.88} & 0.20\pmstd{0.01} & 0.420\pmstd{0.030} \\
\midrule
AgentGuard
& 0.258\pmstd{0.019} & 25.09\pmstd{0.91} & 0.22\pmstd{0.02} & 0.450\pmstd{0.019} \\
AgentForesight
& 0.168\pmstd{0.011} & 24.17\pmstd{0.91} & \second{0.24}\pmstd{0.01} & 0.430\pmstd{0.016} \\
AgentFixer
& 0.171\pmstd{0.017} & 26.69\pmstd{0.80} & 0.22\pmstd{0.01} & \second{0.687}\pmstd{0.027} \\
Wink
& \second{0.264}\pmstd{0.015} & \second{27.56}\pmstd{0.98} & 0.18\pmstd{0.02} & 0.370\pmstd{0.020} \\
\midrule
\rowcolor{sentrybg}
Sentry
& \textbf{0.468}\pmstd{0.017} & \textbf{38.53}\pmstd{0.87} & \textbf{0.30}\pmstd{0.01} & \textbf{0.736}\pmstd{0.019} \\
\metric{\itshape \quad Gain over best baseline}
& \gain{77.3} & \gain{39.8} & \gain{25.0} & \gain{7.1} \\
\bottomrule
\end{tabular}
\end{table}

\paragraph{Sentry improves over both runtime intervention and context evolution baselines.}
Sentry improves on the strongest runtime-intervention baseline for each benchmark by 37\% on average (up to 77\% on WebShop; Table~\ref{tab:main_results_all}) and on ACE, the strongest context-evolution baseline, by 39\% (Table~\ref{tab:self_evolving_baselines}). Adding Sentry to ACE gives the best result on both benchmarks, suggesting that task-level and failure-level knowledge are complementary. Secondary metrics are in Appendix~\ref{app:full_main_results}.

\begin{table}[t]
\centering
\caption{\textbf{Comparison with context-evolution baselines.}
All methods use Qwen3.5-9B and the held-out protocol (\S\ref{protocol}): methods with memory
initialize it on 50 tasks and keep updating it by their own rules while being evaluated on 50 disjoint tasks.
Only held-out tasks are scored, so Base-agent scores differ from Table~\ref{tab:main_results_all}.
Mean $\pm$ std over five runs; \textbf{bold}: best per column; \underline{underline}: second best.
The last row gives Sentry's relative gain over ACE, the strongest context-evolution baseline.}
\label{tab:self_evolving_baselines}
\small
\setlength{\tabcolsep}{9pt}
\renewcommand{\arraystretch}{1.12}
\begin{tabular}{lcc}
\toprule
Method
& \makecell{WebShop\\\metric{Avg.\ reward $\uparrow$}}
& \makecell{Mind2Web Replay\\\metric{Completion $\uparrow$}} \\
\midrule
Base agent    & 0.186\pmstd{0.014} & 0.421\pmstd{0.026} \\
\midrule
Reflexion     & 0.295\pmstd{0.025} & 0.531\pmstd{0.023} \\
ACE           & 0.335\pmstd{0.019} & 0.562\pmstd{0.020} \\
AWM           & 0.270\pmstd{0.021} & 0.470\pmstd{0.018} \\
ReasoningBank & 0.250\pmstd{0.015} & 0.483\pmstd{0.020} \\
\midrule
\rowcolor{sentrybg}
Sentry        & \second{0.495}\pmstd{0.017} & \second{0.729}\pmstd{0.019} \\
\rowcolor{sentrybg}
ACE\,+\,Sentry & \textbf{0.555}\pmstd{0.027} & \textbf{0.789}\pmstd{0.022} \\
\metric{\itshape \quad Sentry vs.\ ACE} & \gain{47.8} & \gain{29.7} \\
\bottomrule
\end{tabular}
\end{table}

\subsection{Intervention effectiveness and human validation}
\label{sec:intervention_effectiveness}

\paragraph{Sentry recovers diverse failure modes.}
\begin{figure}[t]
    \centering
    \includegraphics[width=\linewidth]{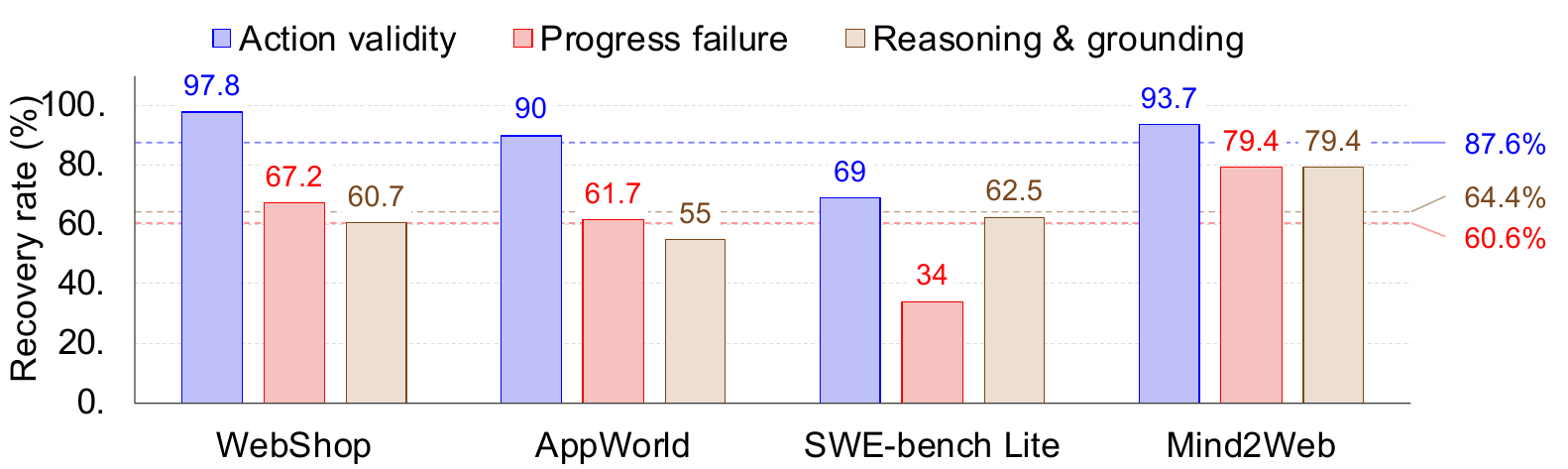}
    \caption{
        \textbf{Local recovery rates by failure category and benchmark.} Recovery rates measure the fraction of detected failures judged resolved within each category and benchmark. Action-validity recovery requires a corrected action accepted by the environment; progress and reasoning–grounding recovery is assessed by Sentry’s verifier after (H=10) reasoning–action–observation cycles.
    }
    \label{fig:failure_resolution}
\end{figure}

Sentry resolves \(81.7\%\) of the 939 detected failures
(Figure~\ref{fig:failure_resolution}). Action-validity failures are the most reliably repaired (\(92.6\%\)) because success is immediate and binary: the corrected action is either accepted or rejected by the environment. Progress (\(64.0\%\)) and reasoning--grounding (\(70.2\%\)) failures are harder because they require correcting the agent's behavior over a longer horizon, rather than fixing a single invalid action. Progress recovery is especially difficult on SWE-bench Lite (\(34.0\%\)), where long repository-level tasks make it harder for the agent to get back on track once it gets stuck.

\paragraph{Soft repair improves local recovery and final task performance.}
To further evaluate Sentry's soft repair, we fork 200 soft-repair intervention
checkpoints across the four benchmarks: from the same agent state, one branch
receives Sentry's soft-repair intervention and the other continues unaided, with
no further interventions in either. Under the same recovery-verification criterion, local recovery
increases from $33.5\%$ without intervention to $60.5\%$
with Sentry. The intervened branch achieves a higher final score in 74 of the 200 pairs, ties in 101, and performs worse in 25. This shows the benefit of Sentry's soft repair to both local recovery and final task performance. Full results are provided in
Appendix~\ref{app:paired_soft_repair}.

\paragraph{Sentry's intervention agrees with human judgements.}  
Because the recovery rates above rely on Sentry's own detector and verifier,
we validate them against human annotations on 200 trajectory windows, including triggered windows and
task-matched non-triggered controls. Human agreement is $86.0\%$ for intervention decisions
(200 windows) and $81.8\%$ for recovery judgments
(99 soft-repair windows); $85.8\%$ of the 113 issued
interventions are judged locally helpful. Inter-annotator agreement is high
(Cohen's \(\kappa=0.82\)--\(0.89\)). Full annotation procedures and
per-benchmark results are provided in Appendix~\ref{app:human_annotation_protocol}.

\subsection{Ablations}
\label{sec:ablations}
\paragraph{All components contribute to Sentry's performance.}
We perform component ablations by removing one Sentry component at a time
(Table~\ref{tab:sentry_component_ablation}); every ablation reduces performance
on both benchmarks. Removing soft repair reduces performance by \(41.9\%\) on
WebShop and \(30.3\%\) on AppWorld. Removing the
intervention-need check yields the largest average relative
drop across the two benchmarks (\(54.7\%\)), highlighting
the importance of selective intervention.
The playbook ablations show another pattern: removing self-evolution, failure-specific
retrieval, and recovery verification reduces performance by
\(31\%\), \(26\%\), and \(20\%\) on average, respectively. Together, these results suggest that Sentry benefits largely from accumulating recovery experience, with further gains from verifying what is learned and retrieving it only for relevant failures. The variant definitions and additional ablation results are provided in
Appendix~\ref{app:full_component_ablation}.

\begin{table}[t]
\centering
\caption{\textbf{Component ablations.}
Each variant removes one component from Sentry, using the online-from-scratch protocol and the same
five runs as Table~\ref{tab:main_results_all}. Entries are mean primary metrics; red values give the
relative drop from Full Sentry. Variant definitions and secondary metrics are in Appendix~\ref{app:full_component_ablation}.}
\label{tab:sentry_component_ablation}
\small
\setlength{\tabcolsep}{6pt}
\renewcommand{\arraystretch}{1.12}
\begin{tabular}{l r@{\hspace{6pt}}l r@{\hspace{6pt}}l}
\toprule
& \multicolumn{2}{c}{WebShop} & \multicolumn{2}{c}{AppWorld} \\
\cmidrule(lr){2-3}\cmidrule(lr){4-5}
Variant
& \multicolumn{2}{c}{\metric{Avg.\ reward $\uparrow$}}
& \multicolumn{2}{c}{\metric{Test pass (\%) $\uparrow$}} \\
\midrule
\rowcolor{sentrybg}
Full Sentry & \textbf{0.468} & & \textbf{38.53} & \\
\midrule
\grouprow{5}{Detection and repair} \\
\quad w/o hard repair              & 0.297 & \drop{36.5} & 31.20 & \drop{19.0} \\
\quad w/o soft repair              & 0.272 & \drop{41.9} & 26.85 & \drop{30.3} \\
\quad w/o intervention-need check  & 0.136 & \drop{70.9} & 23.70 & \drop{38.5} \\
\addlinespace[2pt]
\grouprow{5}{Recovery playbook} \\
\quad w/o self-evolution           & 0.295 & \drop{37.0} & 28.90 & \drop{25.0} \\
\quad w/o failure-specific retrieval & 0.316 & \drop{32.5} & 31.05 & \drop{19.4} \\
\quad w/o recovery verification    & 0.343 & \drop{26.7} & 33.50 & \drop{13.1} \\
\bottomrule
\end{tabular}
\end{table}

\paragraph{Recovery knowledge remains useful on held-out tasks.}
We initialize the playbook on 50 tasks and evaluate
on 50 disjoint tasks
(Table~\ref{tab:playbook_transfer}).
With updates disabled, the frozen playbook improves
WebShop reward from \(0.251\) to \(0.364\)
and Mind2Web Replay completion from \(0.533\)
to \(0.651\), compared with disabling retrieval.
Continuing verified updates further increases
these scores to \(0.495\) and \(0.729\).
These results support reuse of recovery knowledge
across disjoint tasks within the evaluated settings
and the additional benefit of continued adaptation.

\paragraph{Persistent playbook exposure lowers mean performance.}
To examine exposure separately from memory evolution,
we compare two variants using the same frozen
playbook and runtime repair mechanisms.
The exposure variant additionally provides the
complete playbook as persistent background context
to the task agent, while retaining failure-triggered
retrieval and intervention.
This lowers mean WebShop reward from \(0.364\)
to \(0.296\), and mean Mind2Web Replay completion
from \(0.651\) to \(0.599\).
The absolute decreases are \(0.068\) and \(0.052\),
respectively.
This comparison supports keeping the complete
playbook outside the task agent's background context
while providing relevant guidance during interventions.
The detailed control is described in
Appendix~\ref{app:playbook_reuse_exposure}.

\begin{table}[t]
\centering
\caption{\textbf{Reuse and exposure of the recovery playbook on held-out tasks.}
Variants with a playbook start from the same one, initialized on 50 tasks; all methods are evaluated on
the same 50 disjoint tasks (\S\ref{protocol}). Sentry variants keep the detector and both repair
routes and differ only in the playbook settings marked. \emph{Persistent}: the full playbook is also placed
in the task agent's background context. The two frozen variants differ only in this setting; red values
give the change from Frozen playbook. Mean $\pm$ std over five runs; \textbf{bold}: best per column.}
\label{tab:playbook_transfer}
\small
\setlength{\tabcolsep}{5pt}
\renewcommand{\arraystretch}{1.12}
\begin{tabular}{l ccc r@{\hspace{5pt}}l r@{\hspace{5pt}}l}
\toprule
& \multicolumn{3}{c}{Playbook} & \multicolumn{2}{c}{WebShop} & \multicolumn{2}{c}{Mind2Web Replay} \\
\cmidrule(lr){2-4}\cmidrule(lr){5-6}\cmidrule(lr){7-8}
Variant
& \metric{Retrieval} & \metric{Updates} & \metric{Persistent}
& \multicolumn{2}{c}{\metric{Avg.\ reward $\uparrow$}}
& \multicolumn{2}{c}{\metric{Completion $\uparrow$}} \\
\midrule
Base agent
& & & & 0.186\pmstd{0.014} & & 0.421\pmstd{0.026} & \\
\midrule
No external playbook
& \xmark & \xmark & \xmark & 0.251\pmstd{0.017} & & 0.533\pmstd{0.019} & \\
Frozen playbook
& \cmark & \xmark & \xmark & 0.364\pmstd{0.019} & & 0.651\pmstd{0.023} & \\
\quad + persistent exposure
& \cmark & \xmark & \cmark & 0.296\pmstd{0.017} & \dropabs{0.068} & 0.599\pmstd{0.018} & \dropabs{0.052} \\
\rowcolor{sentrybg}
Full Sentry
& \cmark & \cmark & \xmark & \textbf{0.495}\pmstd{0.017} & & \textbf{0.729}\pmstd{0.019} & \\
\bottomrule
\end{tabular}
\end{table}

\paragraph{Playbook retrieval works best when matched to the diagnosed failure.} 
Given that the playbook contains reusable recovery knowledge, we next ablate
how Sentry selects entries for soft repair. As retrieval becomes less specific to the current failure, performance consistently drops. Restricting retrieval only by the broad failure type reduces performance by approximately 11.3\% (relative to Full Sentry, averaged over the two benchmarks); retrieving from the full playbook without filtering reduces it by approximately 18.8\%; and randomly selecting playbook entries reduces it by approximately 28.9\%. This shows that the playbook is most useful when Sentry retrieves experience that matches the failure it has diagnosed.

Performance is less sensitive to the number of retrieved entries: under the same calculation, \(k=3\) and \(k=10\) reduce performance by approximately \(4.5\%\) and \(6.2\%\), respectively, relative to \(k=5\). We therefore use label-filtered retrieval with $k=5$
as the default setting. Full results are in Appendix~\ref{app:retrieval_ablations}.

\paragraph{Sentry is robust to context-window size and verification horizon.}
On WebShop, reward stays within 0.431--0.468 for $W\in\{3,5,7\}$ and 0.458--0.468 for $H\in\{5,10,20\}$, dropping only when the window is too short or too long ($W{=}1$: 0.352; $W{=}15$: 0.384) or verification comes too early ($H{=}3$: 0.432; Appendix~\ref{app:hyperparameter_sensitivity}).

\subsection{Efficiency and robustness}

\paragraph{Sentry's performance gains remain robust across models.}
With the larger GPT-OSS-120B as both the task agent and Sentry's component model, Sentry remains the strongest method, outperforming the strongest runtime-intervention baseline by 51\% on average (95\% on WebShop, 8\% on Mind2Web Replay; Appendix~\ref{app:gptoss120b}). The larger agent still loops and reasons without grounding, and Sentry's isolated failure management continues to improve performance.

\begin{table}[t]
\centering
\caption{\textbf{Token and latency overhead.}
(a) Input plus output tokens, in millions; \emph{Rel.} is each method's total divided by the base agent's.
(b) Change in mean per-task wall-clock time relative to the base agent, in seconds; negative values
mean faster than the base agent.}
\label{tab:cost_speed_summary}
\small
\setlength{\tabcolsep}{6pt}
\renewcommand{\arraystretch}{1.12}
\sisetup{retain-explicit-plus = true}
\begin{tabular}{l
  S[table-format=+2.2] S[table-format=+2.2] S[table-format=+2.2] S[table-format=+2.2]
  S[table-format=2.2] c}
\toprule
Method & {WebShop} & {AppWorld} & {\makecell{SWE-bench\\Lite}} & {\makecell{Mind2Web\\Replay}} & {Total} & {Rel.} \\
\midrule
\grouprow{7}{(a) Tokens, input + output (millions) $\downarrow$} \\
Base agent     &  6.19 &  4.22 & 1.06 & 0.86 & 12.32 & 1.00$\times$ \\
AgentGuard     &  7.53 &  5.39 & 1.34 & 1.01 & 15.26 & 1.24$\times$ \\
AgentForesight & 15.32 & 10.06 & 2.48 & 2.22 & 30.08 & 2.44$\times$ \\
AgentFixer     &  8.20 &  5.61 & 1.41 & 1.11 & 16.33 & 1.33$\times$ \\
Wink           &  6.24 &  4.23 & 1.07 & 0.88 & 12.42 & 1.01$\times$ \\
\rowcolor{sentrybg}
Sentry         &  9.30 &  6.07 & 1.73 & 1.91 & 19.01 & 1.54$\times$ \\
\midrule
\grouprow{7}{(b) Per-task wall-clock time vs.\ base agent (s) $\downarrow$} \\
AgentGuard     &  -3.78 & -26.06 & -14.39 &  +2.70 & & \\
AgentForesight & +22.98 & -34.70 &  -5.35 &  -1.83 & & \\
AgentFixer     & -12.54 & -57.54 & -19.01 & +28.63 & & \\
Wink           &  -1.70 & -47.76 & -15.24 &  -4.25 & & \\
\rowcolor{sentrybg}
Sentry         &  +7.12 & -45.61 & -31.73 & +46.04 & & \\
\bottomrule
\end{tabular}
\end{table}

\paragraph{Sentry adds moderate token usage but can reduce time on long-horizon tasks.}
\label{sec:efficiency_robustness}
Across the four benchmarks, Sentry's token usage is
$1.54\times$ that of the base agent, versus
$2.44\times$ for AgentForesight
(Table~\ref{tab:cost_speed_summary}a).
Wall-clock time moves in both directions: per task, Sentry adds \(7.1\)s on
WebShop and \(46.0\)s on Mind2Web Replay, but removes \(31.7\)s on SWE-bench Lite and
\(45.6\)s on AppWorld (Table~\ref{tab:cost_speed_summary}b). This suggests that on long tasks like SWE-bench, where looping can waste much of the budget, recovering stalled trajectories can save more time than Sentry spends on recovery.

%% file: sections/discussion_final.tex
\section{Conclusion and limitations}
 
\paragraph{Conclusion.}
Failure knowledge is conditional knowledge and should be conditionally exposed.
Sentry puts this principle into practice: it keeps an evolving recovery
playbook outside the agent's context, surfaces a few lessons only when it
diagnoses a matching failure, and admits a lesson only after verifying the
recovery. It outperforms runtime-intervention baselines on four benchmarks and
context-evolution baselines on two, its lessons transfer to held-out tasks, and
exposing the same playbook persistently lowers performance.
 
\paragraph{Limitations and future work.}
Our experiments use Qwen3.5-9B and GPT-OSS-120B; failure patterns may differ
for other models and agent settings. Sentry uses the same model for the task
agent, detector, and verifier; although its judgments largely agree with human
annotators, decoupled detector and verifier models remain to be studied.
Finally, the playbook may accumulate redundant or overly specific entries,
motivating confidence-based admission, merging, and pruning.

%% file: sections/appendix_final.tex
\appendix
\section{Appendix}

\subsection{Additional Related Work}
\label{app:related_work}

\subsubsection{Context Evolution}

As discussed in Section~\ref{sec:background}, context-evolution methods
accumulate reusable experience across tasks and expose it to the task agent
during future executions. Representative approaches include Reflexion,
AWM, ReasoningBank, and ACE, which store experience as reflections,
workflows, reasoning strategies, or evolving playbooks
\citep{shinn2023reflexion,wang2025awm,ouyang2026reasoningbank,zhang2026ace}.
Related approaches such as ExpeL and Voyager similarly reuse prior
experience through retrieved insights or executable skills
\citep{zhao2023expel,wang2023voyager}, while GEPA evolves prompts from
execution feedback~\citep{agrawal2026gepa}.

Sentry differs in the role and activation of the stored knowledge.
Its persistent memory is restricted to failure-recovery experience and
remains outside the task agent's persistent context. Relevant recovery
lessons are introduced only after a matching runtime failure is detected.
This design is complementary to task-level context evolution, as examined
by our ACE+Sentry experiments.

\subsubsection{Runtime Intervention}

Runtime-intervention methods instead operate on the current execution,
monitoring trajectories and providing corrective feedback when intervention
criteria are met. As summarized in the main text, AgentGuard,
AgentForesight, Wink, and AgentFixer use different combinations of runtime
monitoring, failure prediction, validation, and corrective guidance
\citep{koohestani2025agentguard,zhang2026agentforesight,
nanda2026wink,mulian2026agentfixer}.
Earlier feedback-driven methods such as Self-Refine and CRITIC establish
related mechanisms for improving a current attempt through critique or
external feedback
\citep{madaan2023selfrefine,gou2023critic}.
Dubey~\cite{dubey2026realtime} additionally studies lightweight monitoring,
deterministic checks, and rollback-and-rerun repair.

Sentry builds on the runtime-intervention setting but additionally retains
verified recovery experience across tasks. Rather than generating each
intervention only from the current trajectory, Sentry can retrieve recovery
principles learned from earlier successful repairs.

\subsubsection{Harness Adaptation and Failure Diagnosis}

A complementary line of work improves reliability by modifying the harness
that mediates observations, tool use, and execution. Life-Harness,
AutoHarness, Self-Harness, and HarnessFix adapt or synthesize harness
components from observed failures
\citep{xu2026adaptinginterface,lou2026autoharness,
zhang2026selfharness,chen2026failedtrajectories}.
Sentry does not modify its monitoring, routing, or retrieval procedures
during execution; its evolving state is the external recovery knowledge used
by those procedures.

Failure-analysis work instead characterizes where agent executions break
down. MAST studies failure modes in multi-agent systems
\citep{cemri2025mast}, while AgentErrorTaxonomy, AgentErrorBench, and
AgentDebug analyze failures across planning, reasoning, memory, action, and
system-level execution~\citep{zhu2025where}. Sentry uses a narrower,
operational failure categorization: failure types determine whether to
intervene, which repair route to use, and which recovery lessons to retrieve.

\subsection{Detailed Motivating Examples}
\label{app:failure_knowledge_cases}

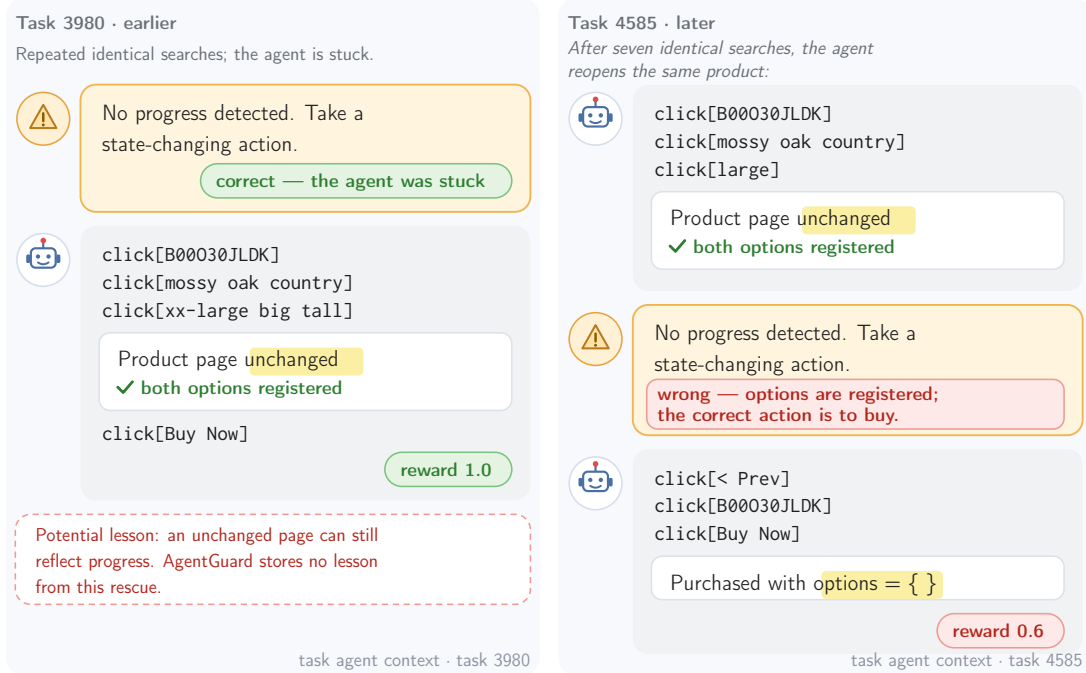
\begin{figure}[t]
    \centering
    \input{figures/detailed_example_figure_appendix.tex}
    \caption{\textbf{Detailed examples underlying Figure~\ref{fig:failure_knowledge_cases}.}
    Trajectories from Qwen3.5-9B on WebShop.
    (a) ACE learns recovery rules from task~6594 and retains them in the agent's context for task~4494. Despite identifying a matching item, the agent continues comparing alternatives and emits no action.
    (b) AgentGuard's fixed warning helps the agent recover from repeated searches in task~3980, but no recovery lesson is stored. In task~4585, the same warning fires after valid option selections leave the page unchanged. The agent then navigates away and purchases the item without those options.}
    \label{fig:failure_knowledge_cases_detailed}
\end{figure}
\clearpage

As shown in Figure~\ref{fig:failure_knowledge_cases_detailed}, we expand the examples in Figure~\ref{fig:failure_knowledge_cases}, showing the recovery rules, agent actions, and task outcomes in greater detail. Both examples use Qwen3.5-9B on WebShop.

\subsection{Experimental Setup Details}
\label{app:experimental_setup}

This section provides additional details for the experimental setup in
Section~\ref{sec:benchmarks_metrics} and Section~\ref{protocol}. We first
describe benchmark-specific details and metrics, then the baseline
implementations, and finally the evaluation protocols used throughout the
experiments.

\subsubsection{Benchmarks and Metrics}
\label{app:Benchmarks and Metrics}
\paragraph{Benchmark details.}

We evaluate on WebShop, AppWorld, SWE-bench Lite, and Mind2Web Replay,
using the benchmark interfaces and primary metrics described in
Section~\ref{sec:benchmarks_metrics}. The main runtime-intervention
comparison uses 100 tasks each for WebShop, AppWorld, and Mind2Web Replay,
and 50 tasks for SWE-bench Lite.

Before sampling, we exclude tasks that cannot be executed in our evaluation
setup because of environment or dependency issues, as well as tasks with
missing required annotations. Tasks are randomly sampled from the remaining
eligible pool. The selected task sets remain fixed across compared methods
and across repeated runs.

WebShop reports average reward as its primary metric. We additionally report
the number of tasks achieving full success (reward \(=1\)) and the number of
tasks in which the agent completes a purchase (\textit{bought}). AppWorld
reports the task test-pass rate: the percentage of evaluation checks passed
per task, averaged across tasks. For SWE-bench Lite, we evaluate generated patches using the official SWE-bench evaluation harness with its Modal backend and report the issue-resolution
accuracy.  Mind2Web Replay reports mean
normalized trajectory completion, defined below.

\paragraph{Mind2Web Replay protocol.}
\textit{Purpose and relation to the original benchmark.}
Mind2Web provides task instructions, annotated action sequences, and recorded
webpage states from real web tasks~\citep{deng2023mind2web}. In the original
evaluation, each reference step is evaluated independently: the model receives
the recorded state and preceding reference history, then predicts the next
action. This measures next-action prediction, but does not evaluate how an
agent recovers after taking an incorrect action.

Because Sentry is designed for runtime failure detection and recovery, we
adapt Mind2Web to allow such recovery while preserving the original tasks,
recorded states, and annotated target actions. In Mind2Web Replay, a rejected
action leaves the agent at the same reference state and returns feedback, so
the agent can revise its next attempt and, when applicable, receive runtime
intervention. The environment advances only after an accepted action.

This preserves the benchmark's core task of selecting the correct action at
each recorded state, while additionally evaluating recovery from execution-time
mistakes. Because Replay allows retries and trajectory-level progress, its
scores are not directly comparable to the original Mind2Web step-wise metrics.

\textit{Data and task construction.}
We use the \texttt{test\_task} split of
\texttt{osunlp/Multimodal-Mind2Web}, corresponding to the Cross-Task
setting.\footnote{\url{https://huggingface.co/datasets/osunlp/Multimodal-Mind2Web}}
We reconstruct each task by grouping action-level records by
\texttt{annotation\_id} and ordering them by
\texttt{target\_action\_index}. We exclude tasks containing a reference
step with no positive candidate element, since such steps cannot be
completed through our candidate-based interface. From the remaining
tasks, we sample 100 for the main evaluation; held-out experiments use
50 initialization tasks and 50 disjoint evaluation tasks.

\textit{Interaction and action evaluation.}
At each reference step, the agent observes the task goal and a candidate
set containing all annotated positive elements and up to 30 negative
elements. Candidates are presented as textual descriptions and assigned
labels such as \texttt{A}, \texttt{B}, and \texttt{AA}. The agent can issue
\texttt{click}, \texttt{type}, or \texttt{select} actions over these labels.

An action is accepted only if it selects an annotated positive candidate,
uses the correct operation, and, for \texttt{TYPE} or \texttt{SELECT},
matches the annotated value. Accepted actions advance to the next recorded
reference state. Rejected actions leave the agent at the current state and
return annotation-derived feedback, allowing another attempt. The same
interface and transition rules are used for all compared methods.

\textit{Scoring.}
For task \(i\), let \(N_i\) be the number of reference steps and \(C_i\)
the number completed in order before termination. We score each task as

\begin{equation}
r_i = \frac{C_i}{N_i},
\end{equation}

and report mean normalized trajectory completion,

\begin{equation}
\overline{r}
=
\frac{1}{M}
\sum_{i=1}^{M}\frac{C_i}{N_i}.
\end{equation}

We also report the number of tasks whose full reference trajectory is
completed. Partial progress is retained when execution terminates before
the final step.

\textit{Scope.}
Mind2Web Replay follows the recorded reference trajectory rather than
executing arbitrary actions on a live website. It therefore evaluates
recovery under a controlled, annotation-conditioned interaction setting.
The benchmark preserves the original task instructions, recorded states,
and annotated target actions, but adds retry-based interaction after
incorrect actions. Its scores should therefore not be compared directly
with the original Mind2Web step-wise metrics.

\paragraph{Additional metric definitions.}

In addition to benchmark-specific task metrics, we report several
intervention-level diagnostics.

\textit{Number of interventions (\# Interv.).}
The total number of interventions issued by a method across evaluated tasks.

\textit{Guided tasks.}
The number of tasks receiving at least one intervention.

\textit{Guidance lift.}
For intervention methods, guidance lift reports the average performance
improvement relative to the base agent on the tasks receiving intervention.
Because methods may intervene on different subsets of tasks, guidance lift
is used as a descriptive intervention-level diagnostic rather than as the
primary metric for comparing methods.

For SWE-bench Lite, where the task-level outcome is binary, we additionally
report the number of guided tasks that are recovered after intervention.

% ============================================================
\subsubsection{Baseline Implementations}
\label{baseline}
% ============================================================

\paragraph{Shared model configuration.}

We evaluate Sentry and the runtime-intervention baselines under a
shared-backbone configuration. In the main experiments, the task agent and
all LLM-based auxiliary components use Qwen3.5-9B, including detection,
auditing, guidance generation, and memory updates where applicable. Across the compared methods, task-agent outputs are limited to
384 generated tokens per call, including any action regeneration
or continuation following an intervention.

We use the same language-model backbone for the task agent and all LLM-based auxiliary components so that differences in performance reflect the intervention methods rather than differences in model capability. Each baseline otherwise retains its method-specific monitoring, intervention, state, and update procedures. Unless stated otherwise, tasks are evaluated independently, with no access to future task trajectories or outcomes.

\paragraph{Runtime intervention baselines.}

\textit{AgentGuard.}
Our AgentGuard implementation follows
\citep{koohestani2025agentguard}. The monitor processes each executed environment transition and maintains
statistics over observed states, actions, transition outcomes, invalid
actions, no-progress transitions, and repeated no-progress actions.
It issues a warning when the invalid-transition rate exceeds \(0.25\), the
no-progress rate exceeds \(0.75\), or the same no-progress action occurs at
least three times consecutively. A new monitor is initialized for each task.
A valid progress-making action resets the repeated no-progress-action streak,
but does not clear the accumulated transition statistics.

\textit{AgentFixer.}
We use the implementation following arXiv:2603.29848v1
\citep{mulian2026agentfixer}.

AgentFixer validates each proposed action using the applicable components of
its validator suite, including unusual-token detection, repetition
detection, output-schema and output-format validation,
information-consistency and action-grounding validation, and
reasoning--action alignment. Findings are assigned
\emph{minor}, \emph{moderate}, or \emph{critical} severity.

When the maximum detected severity is moderate or critical, diagnostic
remediation feedback is returned to the task agent and the action is
regenerated. The validator is stateless and checks each candidate
independently. At most one regeneration is allowed per environment step,
with the retry count resetting at the next step.

\textit{Wink.}
We use the LLM-based trajectory-monitoring and course-correction framework
of Wink~\citep{nanda2026wink}, following arXiv:2602.17037v2.

Under the shared-backbone configuration, Wink's detector uses the same
Qwen3.5-9B model as the task agent. The detector is invoked every five agent
actions and examines the eight most recent trajectory steps for failures
such as specification drift, reasoning problems, and tool-call errors.
Monitoring operates throughout execution on all four benchmarks, including
intermediate execution steps on SWE-bench Lite.

When a failure is detected, the diagnosis and supporting trajectory
evidence are converted into category-specific corrective guidance. The
guidance is appended to the task agent's conversation within
\texttt{<system-reminder>} tags. Each task begins with a fresh trajectory
history. Issuing a reminder does not clear the detector's recent-trajectory
window.

\textit{AgentForesight.}
We use the official AgentForesight inference implementation
\citep{zhang2026agentforesight} at commit
\href{https://github.com/ZBox1005/AgentForesight/commit/af549083fcc6c3eff4ccc558534515e46ca0578e}
{\texttt{af54908}},
following its incremental-prefix prompting and structured-verdict protocol. Before an action is executed, the auditor receives the executed
trajectory prefix together with the proposed action. A valid non-\texttt{SAFE}
verdict blocks the candidate, and its reported failure reason is returned to
the task agent through a domain-specific correction instruction. The
regenerated candidate is then audited again.

Auditor calls are stateless, with all context supplied explicitly. Rejected
candidates are excluded from the executed trajectory; only executed actions
and their resulting observations are included in subsequent trajectory
prefixes.

\paragraph{Context evolution baselines.}
We use the official implementations of Reflexion, ACE, AWM, and
ReasoningBank at the commits listed below, with their method-specific memory
construction, retrieval, and update procedures. All LLM-based components use
the shared task model.

\paragraph{Reflexion.}
We use the official Reflexion implementation~\citep{shinn2023reflexion}
at commit
\href{https://github.com/noahshinn/reflexion/commit/218cf0ef1df84b05ce379dd4a8e47f17766733a0}
{\texttt{218cf0e}}.
Reflexion maintains episodic natural-language reflections derived from task
execution. Its reflection procedure uses the task instruction, execution
trajectory, and feedback to identify failures and formulate lessons for
subsequent tasks. Retrieved reflections are supplied to the task agent before
execution, and new reflections are incorporated after completed tasks.

\paragraph{ACE.}
We use the official ACE implementation~\citep{zhang2026ace}
at commit
\href{https://github.com/ace-agent/ace/commit/82709de050e1db6e6ef2f07bcb0393560b94992a}
{\texttt{82709de}}.
ACE maintains an evolving playbook of reusable task-level advice.
Execution feedback is used to revise and extend the playbook through its
generation, reflection, and curation procedures. Before each task, retrieved
playbook guidance is supplied through the repository's context-construction
procedure. The playbook continues to evolve as tasks are completed.

\paragraph{AWM.}
We use the official AWM implementation~\citep{wang2025awm}
at commit
\href{https://github.com/zorazrw/agent-workflow-memory/commit/8c0ff8cd11d648c8fceb99e4e42f37e3b75381b1}
{\texttt{8c0ff8c}}.
AWM extracts reusable procedural workflows from task trajectories,
abstracting task-specific details into guidance that can be applied to later
tasks. Selected workflows are provided to the task agent before execution.
The workflow memory is updated from completed tasks using the official
workflow-induction procedure.

\paragraph{ReasoningBank.}
We use the official ReasoningBank implementation~\citep{ouyang2026reasoningbank}
at commit
\href{https://github.com/google-research/reasoning-bank/commit/ed80611788292ea739f1effd31f16c53823b8a0d}
{\texttt{ed80611}}.
ReasoningBank distills task trajectories into reusable reasoning insights,
learning from both successful and failed experiences. Relevant entries are
supplied to the task agent before each task. Experience from completed tasks
is subsequently incorporated through the official memory-construction and
update procedures.

\paragraph{ACE+Sentry.}

ACE+Sentry combines task-level context evolution with Sentry's external
runtime failure management. The two methods maintain separate memories and
retain their respective activation and update rules.

ACE maintains its task-level playbook and supplies retrieved entries to the
task agent's initial context before execution. Sentry independently
maintains its recovery playbook outside the task agent's persistent context
and retrieves from it only when a soft-repair intervention is triggered.

During held-out evaluation, ACE updates its playbook after each completed
task according to its standard update procedure. Sentry adds new recovery
entries only after verified soft-repair recoveries. The two playbooks are
not merged.

% ============================================================
\subsection{Implementation details}
\label{app:implementation_details}
 
\paragraph{Serving.}
All model calls, including those made by the task agent, Sentry, and LLM-based
baselines, are served through the Together API.
 
\paragraph{Task selection and repeated runs.}
We use the official benchmark releases and evaluation splits available at the
time of our experiments. Before sampling, we exclude tasks that cannot be
executed in our evaluation setup because of environment or dependency issues,
as well as tasks with missing required annotations. Tasks are randomly sampled
from the remaining eligible pool. The selected task sets remain fixed across
five repeated runs, while the task execution order is randomized for each run.
Under the held-out protocol, the initialization and evaluation sets are fixed
before the repeated runs; execution order is randomized within each set, and
initialization completes before held-out evaluation begins. All updates use
only information available at the time.
 
\paragraph{Sentry configuration.}
In the default configuration, the task agent and all Sentry components use the
same Qwen3.5-9B backbone. The task agent uses a temperature of $0.7$ and a
maximum of 384 generated tokens per call, including action regeneration during
hard repair and continuation after soft-repair guidance. Sentry's auxiliary
calls, including failure detection, recovery verification, and playbook
updating, use a temperature of 0 and a maximum generation length of 4096
tokens. Each component produces structured outputs with fixed fields following
the schemas in Appendix~\ref{prompts}. The detector sees the task objective,
action schema, and recent trajectory window; the verifier sees only the
triggering failure and the subsequent $H$-step recovery trajectory. Neither
observes benchmark rewards or final task success.
 
% \paragraph{Playbook variants in Table~\ref{tab:self_evolving_baselines}.}
% \emph{Sentry, no playbook retrieval} retains Sentry's detector and repair routes
% but disables retrieval. \emph{Sentry, frozen playbook} and \emph{+ persistent
% exposure} start from the same playbook initialized before evaluation, disable
% updates, and retain failure-triggered retrieval; the latter additionally
% places the complete playbook in the task agent's context from the beginning of
% each task. Full Sentry continues adding entries after verified soft-repair
% recoveries.

% ============================================================
\subsubsection{Evaluation Protocols}
% ============================================================
\label{app:evaluation_protocols}

We use the two evaluation protocols defined in
Section~\ref{protocol}: an \emph{online-from-scratch protocol} for the main
runtime-intervention comparison and component ablations, and a
\emph{held-out protocol} for context-evolution comparisons and experiments
involving pre-initialized recovery memory. This section provides additional
implementation details.

\paragraph{Task selection and repeated runs.}
For each experiment, the sampled task set is fixed across all compared
methods and across the five repeated runs. All methods within a comparison use the same task set,
environment interface, task-agent model, and execution limits. Tasks that
cannot be executed in our setup because of environment or dependency
failures, or that lack required annotations, are removed before sampling. All methods use task-agent temperature 0.7, with fresh generation samples and randomized task order across runs on a fixed task set.

\paragraph{Online-from-scratch protocol.}
As described in Section~\ref{protocol}, Sentry starts each run with an empty
recovery playbook. Playbook entries available at any point therefore come
only from verified soft-repair recoveries encountered earlier in that run.
No recovery knowledge from future tasks or outcomes is available.

Component-ablation variants use the same task sequence and initialization
as Full Sentry. Unless an ablation explicitly modifies memory construction,
retrieval, or verification, the remaining Sentry components follow the Full
Sentry procedure.

\paragraph{Held-out protocol.}
The held-out protocol uses 50 initialization tasks followed by 50 disjoint
evaluation tasks, with only the evaluation tasks contributing to reported
scores (Section~\ref{protocol}). During evaluation, methods continue updating
their memories according to their own update procedures unless the experiment
explicitly freezes memory.

For Sentry, updates occur only after verified soft-repair recoveries. For
context-evolution baselines, updates occur according to each method's
task-level memory-update procedure after completed tasks. In all cases,
updates use only information observed up to the current point in the task
sequence.

For the frozen-playbook exposure control, both variants begin evaluation
from the same playbook produced during initialization and disable all further
playbook updates. They use the same task order, detector, repair procedures,
retrieval rule, and recovery verifier. The only difference is whether the
complete frozen playbook is additionally placed in the task agent's
persistent context. Failure-triggered retrieval remains enabled in both
conditions.

\subsection{Additional Main Results}
\label{app:additional_results}

This section provides the complete results underlying the main comparisons
in Section~\ref{sec:main_results}. We first report benchmark-specific
secondary metrics and intervention diagnostics, followed by results with
GPT-OSS-120B and the efficiency analysis.

\subsubsection{Full Benchmark Results}
\label{app:full_main_results}

Tables~\ref{tab:full_webshop_results}--\ref{tab:full_mind2web_results}
report the complete results for the runtime-intervention comparison in
Table~\ref{tab:main_results_all}. In addition to each benchmark's primary
metric, we report benchmark-specific secondary metrics and the
intervention-level diagnostics defined in
Appendix~\ref{app:experimental_setup}.

\begin{table}[!t]
\centering
\setlength{\abovecaptionskip}{\baselineskip}
\setlength{\belowcaptionskip}{\baselineskip}
\caption{
Full WebShop results using Qwen3.5-9B.
}
\small
\setlength{\tabcolsep}{4pt}
\renewcommand{\arraystretch}{1.08}
\begin{tabular}{lrrrrrr}
\toprule
Method
& \makecell{Avg.\\reward}
& \makecell{Full\\succ.}
& Bought
& \makecell{\#\\Interv.}
& \makecell{\# Guided\\tasks}
& \makecell{Guid.\\lift} \\
\midrule
Base agent
& \(0.170 \pm 0.016\)
& 6
& 23
& 0
& 0
& -- \\
AgentGuard
& \(0.258 \pm 0.019\)
& 7
& 37
& 533
& 65
& 0.134 \\
AgentForesight
& \(0.168 \pm 0.011\)
& 4
& 25
& 15
& 14
& -0.018 \\
AgentFixer
& \(0.171 \pm 0.017\)
& 5
& 24
& 236
& 46
& 0.002 \\
Wink
& \(0.264 \pm 0.015\)
& 10
& 33
& 137
& 60
& 0.156 \\
\textbf{Sentry}
& \(\mathbf{0.468 \pm 0.017}\)
& \textbf{14}
& \textbf{79}
& 227
& \textbf{93}
& \textbf{0.320} \\
\bottomrule
\end{tabular}
\label{tab:full_webshop_results}
\end{table}

\begin{table}[t]
\centering
\setlength{\abovecaptionskip}{\baselineskip}
\setlength{\belowcaptionskip}{\baselineskip}
\caption{Full AppWorld results using Qwen3.5-9B. Guidance lift is in
percentage points of test pass rate.}
\small
\setlength{\tabcolsep}{5pt}
\renewcommand{\arraystretch}{1.08}
\begin{tabular}{lrrrr}
\toprule
Method
& \makecell{Test\\pass (\%)}
& \makecell{\#\\Interv.}
& \makecell{\# Guided\\tasks}
& \makecell{Guid.\\lift} \\
\midrule
Base agent
& \(25.59 \pm 0.88\) & 0 & 0 & -- \\
AgentGuard
& \(25.09 \pm 0.91\) & 630 & 54 & -0.93 \\
AgentForesight
& \(24.17 \pm 0.91\) & 25 & 12 & -11.83 \\
AgentFixer
& \(26.69 \pm 0.80\) & 14 & 8 & 13.75 \\
Wink
& \(27.56 \pm 0.98\) & 24 & 14 & 14.07 \\
\textbf{Sentry}
& \(\mathbf{38.53 \pm 0.87}\) & 187 & \textbf{66} & \textbf{19.61} \\
\bottomrule
\end{tabular}
\label{tab:full_appworld_results}
\end{table}

\begin{table}[t]
\centering
\setlength{\abovecaptionskip}{\baselineskip}
\setlength{\belowcaptionskip}{\baselineskip}
\caption{
Full SWE-bench Lite results using Qwen3.5-9B.
}
\small
\setlength{\tabcolsep}{6pt}
\renewcommand{\arraystretch}{1.08}
\begin{tabular}{lrrrr}
\toprule
Method
& Acc.
& \makecell{\#\\Interv.}
& \makecell{\# Guided\\tasks}
& \makecell{\#\\Recovered} \\
\midrule
Base agent
& \(0.20 \pm 0.01\)
& 0
& 0
& -- \\
AgentGuard
& \(0.22 \pm 0.02\)
& 6
& 3
& 1 \\
AgentForesight
& \(0.24 \pm 0.01\)
& 22
& 15
& 3 \\
AgentFixer
& \(0.22 \pm 0.01\)
& 75
& 38
& 4 \\
Wink
& \(0.18 \pm 0.02\)
& 4
& 3
& 3 \\
\textbf{Sentry}
& \(\mathbf{0.30 \pm 0.01}\)
& 90
& 9
& \textbf{5} \\
\bottomrule
\end{tabular}
\label{tab:full_swebench_results}
\end{table}

\begin{table}[t]
\centering
\setlength{\abovecaptionskip}{\baselineskip}
\setlength{\belowcaptionskip}{\baselineskip}
\caption{
Results on Mind2Web Replay using Qwen3.5-9B.
Mean replay completion is the average fraction
of reference steps completed in order.
Fully completed tasks counts reference trajectories
completed in their entirety.
The protocol is specified in
Appendix~\ref{app:Benchmarks and Metrics}.
}
\small
\setlength{\tabcolsep}{5pt}
\renewcommand{\arraystretch}{1.08}
\begin{tabular}{lrrrrr}
\toprule
Method
& \makecell{Mean replay\\completion}
& \makecell{Fully completed\\tasks}
& \makecell{\#\\Interv.}
& \makecell{\# Guided\\tasks}
& \makecell{Guid.\\lift} \\
\midrule
Base agent
& \(0.420 \pm 0.030\)
& 32
& 0
& 0
& -- \\
AgentGuard
& \(0.450 \pm 0.019\)
& 37
& 448
& 94
& 0.032 \\
AgentForesight
& \(0.430 \pm 0.016\)
& 34
& 470
& 93
& 0.011 \\
AgentFixer
& \(0.687 \pm 0.027\)
& 50
& 517
& 70
& 0.382 \\
Wink
& \(0.370 \pm 0.020\)
& 32
& 7
& 6
& -0.840 \\
\textbf{Sentry}
& \(\mathbf{0.736 \pm 0.019}\)
& \textbf{55}
& 435
& 77
& \textbf{0.410} \\
\bottomrule
\end{tabular}
\label{tab:full_mind2web_results}
\end{table}

\subsubsection{Results with GPT-OSS-120B}
\label{app:gptoss120b}

To examine whether the main results depend on the default Qwen3.5-9B
backbone, we repeat the runtime-intervention comparison on WebShop and
Mind2Web Replay using GPT-OSS-120B for both the task agent and the
LLM-based intervention components. The results are reported in
Table~\ref{tab:gptoss120b_results}.

Sentry obtains the highest measured primary task metric on both benchmarks
in this setting. These results provide evidence that the gains observed with
Qwen3.5-9B are not specific to that backbone, although the GPT-OSS-120B
evaluation covers only two of the four benchmarks.
\begin{table}[t]
\centering
\setlength{\abovecaptionskip}{\baselineskip}
\setlength{\belowcaptionskip}{\baselineskip}
\caption{
Results using GPT-OSS-120B on WebShop and Mind2Web Replay.
WebShop reports average reward and full-success/purchase counts;
Mind2Web Replay reports mean normalized trajectory completion and the
number of fully completed tasks.
}
\small
\setlength{\tabcolsep}{3.5pt}
\renewcommand{\arraystretch}{1.08}
\begin{tabular}{lcccccc}
\toprule
Method
& \makecell{WebShop\\reward}
& \makecell{WebShop\\succ./bought}
& \makecell{WebShop\\guid. lift}
& \makecell{Mind2Web \\Replay\\completion}
& \makecell{Mind2Web \\Replay\\completed tasks}
& \makecell{Mind2Web \\Replay\\guid. lift} \\
\midrule

Base agent
& 0.192
& 7/28
& --
& 0.550
& 43
& -- \\

AgentGuard
& 0.246
& 8/35
& 0.092
& 0.740
& 68
& 0.194 \\

AgentForesight
& 0.091
& 4/12
& -0.186
& 0.582
& 52
& 0.033 \\

AgentFixer
& 0.273
& 5/44
& 0.112
& 0.753
& 62
& \textbf{0.328} \\

Wink
& 0.197
& 7/33
& 0.008
& 0.580
& 47
& 0.111 \\

\textbf{Sentry}
& \textbf{0.531}
& \textbf{12/92}
& \textbf{0.357}
& \textbf{0.815}
& \textbf{75}
& 0.294 \\

\bottomrule
\end{tabular}
\label{tab:gptoss120b_results}
\end{table}

% ============================================================
\subsection{Additional Intervention Analysis and Ablations}
\label{app:intervention_analysis}
% ============================================================

This section provides the full intervention-level analyses underlying the
results in Section~\ref{sec:intervention_effectiveness}. In the main text,
we show that Sentry recovers diverse failure modes and that soft repair
improves both local recovery and final task performance. Here, we provide
the complete paired soft-repair results and benchmark-level failure-recovery
breakdown.

% ============================================================
\subsubsection{Paired Soft-Repair Evaluation}
\label{app:paired_soft_repair}
% ============================================================

As described in Section~\ref{sec:intervention_effectiveness}, we fork 200
soft-repair checkpoints across the four benchmarks. At each checkpoint, both
branches begin from the same agent state and trajectory history: one receives
Sentry's soft-repair intervention and the other continues without
intervention. No further interventions are allowed in either branch.

Local recovery is evaluated after the same \(H=10\) horizon used by
Sentry's recovery verifier, and both branches are then continued to
termination for the benchmark-specific final task metric.

Table~\ref{tab:counterfactual_recovery} reports the complete results behind
the main-text comparison. Because both branches share the same
pre-intervention state, this experiment directly compares continuations with
and without soft repair.

\begin{table}[t]
\centering
\setlength{\abovecaptionskip}{\baselineskip}
\setlength{\belowcaptionskip}{\baselineskip}
\caption{
Paired counterfactual evaluation at 200 soft-repair checkpoints.
Local recovery is judged after \(H=10\) steps. The final row counts
pairs in which each branch obtains a higher final score under the task-level metric defined in
Section~\ref{sec:benchmarks_metrics}; the remaining 101 pairs are tied.
}
\small
\setlength{\tabcolsep}{6pt}
\renewcommand{\arraystretch}{1.1}
\begin{tabular}{lcc}
\toprule
& \textbf{No intervention} & \textbf{Sentry} \\
\midrule
Local recovery (of 200)
& 67 (33.5\%) & \textbf{121 (60.5\%)} \\
Recovered only in this branch
& 12 & \textbf{66} \\
Higher final task score
& 25 & \textbf{74} \\
\bottomrule
\end{tabular}
\label{tab:counterfactual_recovery}
\end{table}

\subsubsection{Component Ablations}
\label{app:full_component_ablation}

We provide the full component-ablation results on WebShop and AppWorld,
expanding the compact comparison reported in
Table~\ref{tab:sentry_component_ablation} in the main text.
Tables~\ref{tab:full_sentry_component_ablation_webshop}
and~\ref{tab:full_sentry_component_ablation_appworld}
report the complete benchmark-specific results, including secondary
metrics and intervention-level diagnostics.

All variants follow the online-from-scratch protocol in
Appendix~\ref{app:evaluation_protocols} and use the same task sets and model
configuration as Full Sentry. Each variant removes or modifies one component
while leaving the remaining procedure unchanged unless the removed component
directly affects downstream operations.

\paragraph{Without hard repair.}
We disable Sentry's action-correction procedure for schema-invalid actions.
When an invalid action is detected, Sentry does not request a corrected
replacement action. The soft-repair route, including failure detection,
playbook retrieval, recovery verification, and playbook updates, remains
enabled.

\paragraph{Without soft repair.}
We disable corrective guidance for progress and reasoning--grounding
failures. Action-validity checking and hard repair remain enabled. Because
Full Sentry adds playbook entries only from verified soft-repair episodes,
this variant also removes the source of new recovery entries.

\paragraph{Without intervention-need check.}
We remove the binary gate that determines whether soft-repair guidance should
be issued. For valid actions, guidance is generated at every scheduled
monitoring point rather than only after the detector identifies a failure.
The remaining repair, verification, and playbook-update procedures are
unchanged.

\paragraph{Without failure-specific retrieval.}
We retain the evolving playbook but remove conditioning on the diagnosed
failure type and fine-grained retrieval labels. Entries are instead selected
without using the current failure diagnosis. Failure detection, soft repair,
recovery verification, and playbook updates remain enabled.

\paragraph{Without playbook self-evolution.}
We disable all playbook updates. Under the online-from-scratch protocol, the
playbook therefore remains empty throughout evaluation. Failure detection and
both repair routes remain enabled, but soft repair relies only on the current
diagnosis and general repair instructions.

\paragraph{Without recovery verification.}
We remove the resolved/unresolved recovery judgment as a prerequisite for
playbook admission. Soft-repair episodes are added to the playbook after the
same observation horizon without requiring a verified recovery. The repair
and retrieval procedures otherwise remain unchanged.

Across both benchmarks, every component ablation reduces the primary task
metric relative to Full Sentry. These results are consistent with
complementary contributions from selective intervention, soft and hard
repair, recovery-memory evolution, failure-conditioned retrieval, and
verification before memory admission.

\begin{table}[t]
\centering
\setlength{\abovecaptionskip}{\baselineskip}
\setlength{\belowcaptionskip}{\baselineskip}
\caption{
Component-ablation results on WebShop. Each variant removes one component from
Sentry. Downward arrows indicate drops relative to Full Sentry for the primary
task-level metric. Guidance lift is computed over guided tasks, following the
definition in Appendix~\ref{app:Benchmarks and Metrics}.
}
\small
\setlength{\tabcolsep}{5pt}
\renewcommand{\arraystretch}{1.08}
\begin{tabular}{@{}lrrrrrr@{}}
\toprule
\textbf{Variant}
& \makecell{\textbf{Avg.}\\\textbf{reward}}
& \makecell{\textbf{Full}\\\textbf{succ.}}
& \textbf{Bought}
& \makecell{\textbf{\#}\\\textbf{Interv.}}
& \makecell{\textbf{\# Guided}\\\textbf{tasks}}
& \makecell{\textbf{Guid.}\\\textbf{lift}} \\
\midrule
\textbf{Full Sentry}
& \(\mathbf{0.468}\)
& \(\mathbf{14}\)
& \(\mathbf{79}\)
& 227
& 93
& \(\mathbf{0.320}\) \\

w/o hard repair
& \(0.297\) {\footnotesize \((\downarrow 0.171)\)}
& 5
& 51
& 59
& 53
& 0.240 \\

w/o soft repair
& \(0.272\) {\footnotesize \((\downarrow 0.196)\)}
& 9
& 63
& 159
& 75
& 0.136 \\

w/o playbook failure-specific retrieval
& \(0.316\) {\footnotesize \((\downarrow 0.152)\)}
& 6
& 45
& 102
& 93
& 0.156 \\

w/o playbook self-evolution
& \(0.295\) {\footnotesize \((\downarrow 0.173)\)}
& 4
& 38
& 198
& 89
& 0.140 \\

w/o intervention-need check
& \(0.136\) {\footnotesize \((\downarrow 0.332)\)}
& 1
& 19
& 347
& 99
& \(-0.035\) \\

w/o verification of recovery success
& \(0.343\) {\footnotesize \((\downarrow 0.125)\)}
& 8     
& 55     
& 214   
& 91     
& 0.190
\\
\bottomrule
\end{tabular}
\label{tab:full_sentry_component_ablation_webshop}
\end{table}

\begin{table}[t]
\centering
\setlength{\abovecaptionskip}{\baselineskip}
\setlength{\belowcaptionskip}{\baselineskip}
\caption{
Component-ablation results on AppWorld. Each variant removes one component
from Sentry; arrows give the drop in test pass rate relative to Full
Sentry. Guidance lift is in percentage points and follows the definition
in Appendix~\ref{app:Benchmarks and Metrics}.
}
\small
\setlength{\tabcolsep}{6pt}
\renewcommand{\arraystretch}{1.08}
\begin{tabular}{@{}lrrrr@{}}
\toprule
\textbf{Variant}
& \makecell{\textbf{Test}\\\textbf{pass (\%)}}
& \makecell{\textbf{\#}\\\textbf{Interv.}}
& \makecell{\textbf{\# Guided}\\\textbf{tasks}}
& \makecell{\textbf{Guid.}\\\textbf{lift}} \\
\midrule
\textbf{Full Sentry}
& \(\mathbf{38.53}\) & 187 & 66 & \(\mathbf{19.61}\) \\
w/o hard repair
& \(31.20\) {\footnotesize \((\downarrow 7.33)\)} & 67 & 44 & 12.75 \\
w/o soft repair
& \(26.85\) {\footnotesize \((\downarrow 11.68)\)} & 120 & 54 & 2.33 \\
w/o playbook failure-specific retrieval
& \(31.05\) {\footnotesize \((\downarrow 7.48)\)} & 118 & 66 & 8.27 \\
w/o playbook self-evolution
& \(28.90\) {\footnotesize \((\downarrow 9.63)\)} & 166 & 62 & 5.34 \\
w/o intervention-need check
& \(23.70\) {\footnotesize \((\downarrow 14.83)\)} & 310 & 92 & \(-2.05\) \\
w/o verification of recovery success
& \(33.50\) {\footnotesize \((\downarrow 5.03)\)} & 178 & 64 & 12.36 \\
\bottomrule
\end{tabular}
\label{tab:full_sentry_component_ablation_appworld}
\end{table}

Full Sentry achieves a higher primary task metric than every ablated variant
on both WebShop and AppWorld. The detailed results in
Tables~\ref{tab:full_sentry_component_ablation_webshop}
and~\ref{tab:full_sentry_component_ablation_appworld}
are consistent with the main-text analysis: removing selective intervention,
soft or hard repair, playbook evolution, failure-conditioned retrieval, or
recovery verification reduces performance.

\subsubsection{Held-Out Playbook Controls}
\label{app:playbook_reuse_exposure}

The held-out playbook results are summarized in
Section~\ref{sec:ablations}. Here, we clarify the construction of the
variants used in that comparison.

All variants use the same 50 initialization tasks and 50 disjoint held-out
evaluation tasks. They differ only in how the initialized playbook is used
during evaluation.

\textit{No external playbook} retains Sentry's failure detection and repair
mechanisms but disables playbook retrieval.
\textit{Frozen playbook} starts evaluation from the playbook produced during
initialization, allows failure-triggered retrieval from that playbook, and
disables further playbook updates.
\textit{Frozen playbook + persistent exposure} uses the same frozen
playbook and retrieval procedure, while additionally placing the complete
playbook in the task agent's background context throughout execution.
\textit{Full Sentry} starts from the same initialized playbook and continues
adding entries after verified soft-repair recoveries during held-out
evaluation.

For the frozen-playbook comparison, the two conditions use the same
task-agent model, task order, environment, failure detector, repair
procedures, retrieval rule, and recovery verifier. The playbook remains
frozen throughout evaluation, and the only intentional difference is that
the persistent-exposure variant additionally places the complete playbook
in the task agent's background context.

These variants jointly examine reuse of previously learned recovery
knowledge, continued playbook evolution, and persistent versus
failure-triggered exposure of the playbook.

\subsubsection{Retrieval Ablations}
\label{app:retrieval_ablations}

Table~\ref{tab:retrieval_ablation} presents the full results summarized in
Section~\ref{sec:ablations}. We clarify the retrieval variants used
in that comparison.

All variants use the same held-out evaluation setup and differ only in how
playbook entries are selected for soft repair. \textit{Random retrieval}
samples entries without conditioning on the current failure.
\textit{Unfiltered retrieval} selects from the full playbook without using
the diagnosed failure type or fine-grained labels.
\textit{Failure-type-only retrieval} restricts selection to entries with the
same broad failure type.
\textit{Label-filtered retrieval} additionally requires overlap with the
fine-grained labels assigned to the current diagnosis.

For label-filtered retrieval, \(k\) denotes the maximum number of playbook
entries included in a soft-repair prompt. We evaluate
\(k \in \{3,5,10\}\); \(k=5\) is the default Full Sentry setting.
All other Sentry components are held fixed across these variants.
\begin{table}[t]
\centering
\setlength{\abovecaptionskip}{\baselineskip}
\setlength{\belowcaptionskip}{\baselineskip}
\caption{
Playbook retrieval ablations on held-out WebShop and Mind2Web Replay tasks.
Only the soft-repair retrieval rule is varied; all other Sentry components
are fixed. We report mean \(\pm\) standard deviation over five repeated runs.
Label-filtered retrieval with \(k=5\) is the Full Sentry configuration and
uses the same runs as Table~\ref{tab:self_evolving_baselines} (\(0.495\) and
\(0.729\) at three decimals); relative decreases in the text are computed
against those means.
}
\small
\setlength{\tabcolsep}{5pt}
\renewcommand{\arraystretch}{1.08}
\begin{tabular}{lcc}
\toprule
Retrieval variant
& \makecell{WebShop\\avg. reward}
& \makecell{Mind2Web Replay\\completion} \\\\
\midrule

Random retrieval
& \(0.33 \pm 0.03\)
& \(0.55 \pm 0.02\) \\

Unfiltered retrieval
& \(0.39 \pm 0.01\)
& \(0.61 \pm 0.02\) \\

Failure-type-only retrieval
& \(0.43 \pm 0.01\)
& \(0.66 \pm 0.01\) \\
Label-filtered, \(k=3\)
& \(0.47 \pm 0.01\)
& \(0.70 \pm 0.01\) \\

\textbf{Label-filtered, \(k=5\)}
& \(\mathbf{0.50 \pm 0.02}\)
& \(\mathbf{0.73 \pm 0.02}\) \\

Label-filtered, \(k=10\)
& \(0.46 \pm 0.02\)
& \(0.69 \pm 0.02\) \\
\bottomrule
\end{tabular}
\label{tab:retrieval_ablation}
\end{table}

\subsubsection{Hyperparameter Sensitivity}
\label{app:hyperparameter_sensitivity}

The hyperparameter sensitivity results are summarized in
Section~\ref{sec:ablations}. Here, we provide the full results and clarify
the two parameters varied in these experiments.

\paragraph{Context-window size W.}
The detector observes the most recent \(W\)
reasoning--action--observation cycles when deciding whether a failure has
occurred. We evaluate \(W \in \{1,3,5,7,10,15\}\) on WebShop, while keeping
all other Sentry components and settings fixed. The full results, including
intervention and guidance statistics, are reported in Table~\ref{tab:webshop_context_window}.
The default configuration uses \(W=5\).

\paragraph{Recovery-verification horizon H.}
For soft repair, the recovery verifier observes \(H\) subsequent
reasoning--action--observation cycles before judging whether the triggering
failure has been resolved. We evaluate \(H \in \{3,5,10,20\}\) on WebShop,
with all other settings fixed. The full results include both task
performance and recovery-verifier accuracy, reported in Table~\ref{tab:horizon_sensitivity}. The default configuration uses
\(H=10\).

\begin{table}[t]
\centering
\setlength{\abovecaptionskip}{\baselineskip}
\setlength{\belowcaptionskip}{\baselineskip}
\caption{
Effect of Sentry context-window size on WebShop performance.
}
\small
\setlength{\tabcolsep}{6pt}
\renewcommand{\arraystretch}{1.08}
\begin{tabular}{lrrrrrr}
\toprule
\makecell{Context\\window}
& \makecell{Avg.\\reward}
& \makecell{Full\\succ.}
& Bought
& \makecell{\#\\Interv.}
& \makecell{\# Guided\\tasks}
& \makecell{Guid.\\lift} \\
\midrule
1
& 0.352
& 8
& 60
& 165
& 78
& 0.233 \\
3
& 0.431
& 12
& 73
& 205
& 88
& 0.296 \\
\textbf{5}
& \textbf{0.468}
& \textbf{14}
& \textbf{79}
& \textbf{227}
& \textbf{93}
& \textbf{0.320} \\
7
& 0.456
& 13
& 78
& 238
& 94
& 0.304 \\
10
& 0.421
& 12
& 72
& 255
& 95
& 0.264 \\
15
& 0.384
& 10
& 66
& 275
& 97
& 0.220 \\
\bottomrule
\end{tabular}
\label{tab:webshop_context_window}
\end{table}

\begin{table}[t]
    \centering
    \setlength{\abovecaptionskip}{\baselineskip}
    \setlength{\belowcaptionskip}{\baselineskip}
    \caption{
        Sensitivity to the soft-repair verification horizon $H$ on
        WebShop. Performance and recovery-judge accuracy are stable across
        moderate verification horizons, while verifying too early reduces
        both end-to-end task performance and agreement with human recovery
        judgments.
    }
    \small
    \begin{tabular}{c c c c}
        \toprule
        $H$
        & Avg. reward $\uparrow$
        & Guidance lift $\uparrow$
        & Recovery-judge acc. (\%) $\uparrow$ \\
        \midrule
        3  & 0.432 & 0.283 & 76.0 \\
        5  & 0.458 & 0.307 & 82.0 \\
        \textbf{10}
           & \textbf{0.468}
           & \textbf{0.320}
           & \textbf{84.0} \\
        20 & 0.461 & 0.312 & 83.0 \\
        \bottomrule
    \end{tabular}
    \label{tab:horizon_sensitivity}
\end{table}

\subsection{Human Evaluation}
\label{app:human_annotation_protocol}

This appendix gives the full protocol and results for the human audit
summarized in Section~\ref{sec:intervention_effectiveness}. The audit
checks three of Sentry's runtime judgments against human assessment:
the detector's decision to intervene, the verifier's decision that the
triggering failure has been resolved, and the local helpfulness of the
intervention text.

\paragraph{Sampling.}
We sample 200 trajectory windows from completed Sentry runs, 50 from
each benchmark. Each window contains the recent trajectory seen by the
detector and, when Sentry intervened, the intervention text and the
subsequent $H$-step recovery trajectory. The sample contains 113
windows in which Sentry issued an intervention (99 soft repairs and 14
hard repairs, covering Progress, Reasoning \& Grounding, and Action
Validity failures) and 87 control windows in which it did not. Control
windows are matched to the triggered windows by task: they are drawn
from the same tasks and runs, at steps where the detector did not fire,
so that triggered and control windows share the same task distribution.
Including controls lets intervention accuracy reflect both false alarms
and missed failures rather than false alarms alone.

\paragraph{Annotation procedure.}
Each window is independently labeled by two members of the research
team using the rubric in Table~\ref{tab:human_annotation_rubric}.
The two annotators then discuss any disagreements to reach consensus,
and the resulting consensus labels are used for all Sentry-vs-human
comparisons. Sentry's trigger decision is withheld during the first annotation
stage, and its verifier verdict is withheld throughout annotation. Each window is presented in two stages. In the first stage,
annotators see only the pre-intervention trajectory and assign the
\textit{intervention needed} label, without knowing whether Sentry
intervened at that point; triggered and control windows are
indistinguishable at this stage. In the second stage, for triggered
windows only, the intervention text and the subsequent $H$-step
continuation are revealed, and annotators assign the \textit{recovered}
label from the continuation (has the triggering failure pattern
stopped, with the agent resuming valid, task-relevant progress?) and
the \textit{helpful} label from the intervention text and its local
failure context (does the guidance address the observed failure and
give task-relevant advice, regardless of whether the agent ultimately
follows it?). Annotators assign the \textit{recovered} label based on the
continuation without access to the verifier's resolved/unresolved
verdict.

\begin{table}[t]
\centering
\setlength{\abovecaptionskip}{\baselineskip}
\setlength{\belowcaptionskip}{\baselineskip}
\caption{Human annotation rubric for auditing Sentry trajectory windows.}
\small
\setlength{\tabcolsep}{5pt}
\renewcommand{\arraystretch}{1.12}
\begin{tabular}{p{0.25\linewidth}p{0.68\linewidth}}
\toprule
\textbf{Label} & \textbf{Criterion} \\
\midrule
Intervention needed
& The agent shows an invalid action, repeated no-progress behavior,
poor grounding, or another failure that requires intervention. \\
Recovered
& After Sentry's intervention, the agent stops the triggering failure
pattern and resumes valid, task-relevant progress. \\
Helpful
& The intervention text addresses the observed local failure and gives
task-relevant guidance without unsupported assumptions or objective
drift. \\
\bottomrule
\end{tabular}
\label{tab:human_annotation_rubric}
\end{table}

\paragraph{Metrics.}
We compare Sentry's decisions with the consensus labels using three
metrics, each over a different denominator. \textit{Intervention
accuracy} is computed over all 200 windows: the fraction in which
Sentry's trigger/no-trigger decision matches the intervention-needed
label. \textit{Recovery-judge accuracy} is computed over the 99
soft-repair windows: the fraction in which the verifier's
resolved/unresolved output matches the recovered label. The 14
hard-repair windows are excluded from this metric because hard repair
is verified by execution---the corrected action is either accepted by
the environment or not---rather than by the LLM verifier, so there is
no verifier judgment to audit. \textit{Helpful rate} is computed over
all 113 issued interventions: the fraction judged locally helpful.

\paragraph{Results.}
Table~\ref{tab:human_validation} reports results per benchmark.
Overall, Sentry reaches 86.0\% intervention accuracy, 81.8\%
recovery-judge accuracy, and an 85.8\% helpful rate. Agreement is
highest on Mind2Web Replay and lowest on SWE-bench Lite, and recovery-judge
accuracy is the lowest of the three metrics on every benchmark,
consistent with recovery being the hardest judgment: it requires
assessing a multi-step continuation rather than a single state or a
single message. These results indicate that Sentry usually intervenes
on failures that humans also identify, that its recovery verdicts
generally agree with human assessment, and that the guidance it issues
is mostly judged useful.

\begin{table}[t]
\centering
\setlength{\abovecaptionskip}{\baselineskip}
\setlength{\belowcaptionskip}{\baselineskip}
\caption{
Human validation of Sentry's intervention decisions, recovery
judgments, and intervention helpfulness. $N$ is the number of audited
windows per benchmark. Intervention accuracy is computed over all
windows, recovery-judge accuracy over soft-repair windows (99 in
total), and helpful rate over issued interventions (113 in total). All
values except $N$ are percentages.
}
\small
\setlength{\tabcolsep}{5pt}
\renewcommand{\arraystretch}{1.08}
\begin{tabular}{lcccc}
\toprule
Benchmark
& $N$
& \makecell{Intervention\\acc.}
& \makecell{Recovery-judge\\acc.}
& \makecell{Helpful\\rate} \\
\midrule
WebShop
& 50 & 86.0 & 84.0 & 86.2 \\
AppWorld
& 50 & 86.0 & 81.0 & 85.2 \\
SWE-bench Lite
& 50 & 82.0 & 76.9 & 82.1 \\
Mind2Web Replay
& 50 & 90.0 & 85.2 & 89.7 \\
\midrule
Overall
& 200 & 86.0 & 81.8 & 85.8 \\
\bottomrule
\end{tabular}
\label{tab:human_validation}
\end{table}

\paragraph{Inter-annotator agreement.}
Table~\ref{tab:human_agreement_appendix} reports raw percent agreement
and Cohen's $\kappa$ between the two annotators, computed from their
independently assigned binary labels before any discussion to resolve
disagreements. Agreement is high on all three labels ($\kappa$ between
0.82 and 0.89), indicating consistent application of the annotation
rubric prior to consensus discussion.

\begin{table}[t]
\centering
\setlength{\abovecaptionskip}{\baselineskip}
\setlength{\belowcaptionskip}{\baselineskip}
\caption{Inter-annotator agreement before consensus discussion for the human audit.}
\small
\setlength{\tabcolsep}{6pt}
\renewcommand{\arraystretch}{1.08}
\begin{tabular}{lccc}
\toprule
\textbf{Label}
& \textbf{\(N\)}
& \makecell{\textbf{Percent}\\\textbf{agreement}}
& \textbf{Cohen's \(\kappa\)} \\
\midrule
Intervention needed
& 200
& 94.5
& 0.89 \\
Recovered
& 99
& 90.9
& 0.82 \\
Helpful
& 113
& 95.6
& 0.82 \\
\bottomrule
\end{tabular}
\label{tab:human_agreement_appendix}
\end{table}

\subsection{Sentry Implementation Details}

\subsubsection{Failure category and retrieval label definitions}
\label{app:retrieval_labels}

In this section, we provide the detailed definitions and examples for the major
failure categories as well as the retrieval labels.

\paragraph{Action validity failures.}
Action Validity Failures occur when the agent's action is not
executable under the environment schema. These failures include malformed tool
calls, missing arguments, invalid parameter types, unsupported tools, and
invalid targets or controls.

\paragraph{Progress failures.}
Progress Failures occur when the agent's recent trajectory does not move the task toward completion. These failures indicate that the
agent is stuck, repeating unproductive behavior, exploring without making useful
progress, or failing to terminate at the right time. We use the following four
retrieval labels:

\begin{itemize}[leftmargin=1.4em]
    \item \textbf{Repetition or looping.}
    The agent repeats identical reasoning steps, tool calls,
    queries, or actions without gaining new information.

    \item \textbf{Planning stalls.}
    The agent recognizes the task objective but fails to convert it into a
    concrete next step. A typical example is that the agent generates plans but does not take useful actions.

    \item \textbf{Over-exploration.}
    The agent keeps gathering information even after it has enough evidence to
    make progress toward the task goal.

    \item \textbf{Termination miscalibration.}
    The agent stops too early before satisfying the task objective, or continues
    acting after the task is already effectively complete.
\end{itemize}

\paragraph{Reasoning \& grounding failures.}
Reasoning \& Grounding Failures occur when the agent's claims, plans, or
actions are not supported by its observations. These failures
indicate that the agent is acting on unsupported assumptions, drifting from the
original objective, or choosing actions that do not follow from its
reasoning. We use the following three retrieval labels:

\begin{itemize}[leftmargin=1.4em]
    \item \textbf{Hallucination.}
    The agent introduces unsupported facts, assumptions, tool results, environment states, or task constraints that do not appear in the
    trajectory.

    \item \textbf{Objective drift.}
    The agent shifts away from the user's original objective.

    \item \textbf{Reasoning-action mismatch.}
    The agent's action does not follow from its own reasoning or from the
    current observation, such as choosing an action that contradicts its stated
    plan.
\end{itemize}

\subsubsection{Retrieval and update rules}
\label{app:playbook_retrieval}

This appendix specifies the retrieval rule used by Sentry's external playbook.
Each playbook entry is stored as a structured record containing a failure type,
retrieval labels, a trigger pattern, a repair principle, and an insertion step.
Sentry uses a deterministic label-filtered retrieval rule rather than embedding-based
semantic search, making the retrieval behavior inspectable.

For a soft-repair failure, Sentry retrieves playbook entries in three steps. First, it keeps only entries with the same failure type and at least one retrieval label in common with the current failure. Second, it ranks the remaining entries by the number of overlapping labels, using recency only to break ties. Third, it inserts up to \(k=5\) entries into the soft-repair prompt. When more than five entries are available, Sentry greedily selects the highest-ranked entries, while limiting entries with the same primary retrieval label to at most two whenever possible. This diversity cap prevents guidance from being dominated by a single repeated failure subtype. If no playbook entry shares a retrieval label with the diagnosis, Sentry
falls back to the \(k\) most recent entries of the same failure type.

The playbook is updated only after recovery verification. After a soft repair,
Sentry waits \(H\) reasoning--action--observation cycles and invokes the
verifier. If the local failure is judged as resolved, the successful recovery is
summarized into a new playbook entry. Although the
physical playbook can grow over time, the inference-time guidance remains bounded
by top-\(k\) retrieval, so the amount of injected guidance does not grow linearly
with the playbook size.

\subsubsection{Sentry prompt templates}
\label{prompts}

This section provides the prompt skeletons used by Sentry. We show the shared
structure rather than benchmark-specific schemas.

\begin{tcolorbox}[
  enhanced,
  breakable,
  title={Sentry detector prompt skeleton},
  colback=gray!3,
  colframe=black!55,
  coltitle=white,
  colbacktitle=black!55,
  fonttitle=\bfseries,
  boxrule=0.5pt,
  arc=2pt,
  left=4pt,
  right=4pt,
  top=4pt,
  bottom=4pt
]
\textcolor{black!70}{\texttt{[SENTRY DETECTOR]}} \\[2pt]

\textbf{Role.} \\
You are Sentry, an external runtime failure manager for an LLM task agent.
Inspect the recent trajectory and decide whether the agent needs intervention.
Do not judge final task success; judge whether the current local trajectory
contains an execution-time failure. \\[3pt]

\textbf{Task objective.} \\
\{\texttt{task\_objective}\} \\[3pt]

\textbf{Recent trajectory window.} \\
\{\texttt{recent\_trajectory}\} \\[3pt]

\textbf{Required action schema.} \\
\{\texttt{schema\_description}\} \\[3pt]

\textbf{Failure taxonomy.} \\
Action Validity: the proposed action is not executable under the environment
schema. \\
Progress Failure: the recent trajectory repeats, stalls, over-explores, or
terminates at the wrong time. \\
Reasoning \& Grounding Failure: the agent introduces unsupported assumptions,
drifts from the objective, or takes an action that does not follow from the
reasoning or observation. \\[3pt]

\textbf{Output format.} \\
Return whether intervention is needed, the failure type, retrieval labels,
local evidence, and a concise diagnosis.
\end{tcolorbox}

\begin{tcolorbox}[
  enhanced,
  breakable,
  title={Sentry hard repair prompt skeleton},
  colback=gray!3,
  colframe=black!55,
  coltitle=white,
  colbacktitle=black!55,
  fonttitle=\bfseries,
  boxrule=0.5pt,
  arc=2pt,
  left=4pt,
  right=4pt,
  top=4pt,
  bottom=4pt
]
\textcolor{black!70}{\texttt{[SENTRY HARD REPAIR]}} \\[2pt]

\textbf{Detected issue.} \\
The previous action is invalid or not executable under the required schema.
Rewrite the action into one valid action that follows the same task objective
and the current observation. \\[3pt]

\textbf{Task objective.} \\
\{\texttt{task\_objective}\} \\[3pt]

\textbf{Recent trajectory.} \\
\{\texttt{recent\_trajectory}\} \\[3pt]

\textbf{Invalid action.} \\
\{\texttt{invalid\_action}\} \\[3pt]

\textbf{Schema error.} \\
\{\texttt{schema\_error}\} \\[3pt]

\textbf{Required schema.} \\
\{\texttt{schema\_description}\} \\[3pt]

\textbf{Continuation instruction.} \\
Output only the corrected action in the required action format.
\end{tcolorbox}

\begin{tcolorbox}[
  enhanced,
  breakable,
  title={Sentry soft repair prompt skeleton},
  colback=gray!3,
  colframe=black!55,
  coltitle=white,
  colbacktitle=black!55,
  fonttitle=\bfseries,
  boxrule=0.5pt,
  arc=2pt,
  left=4pt,
  right=4pt,
  top=4pt,
  bottom=4pt
]
\textcolor{black!70}{\texttt{[SENTRY SOFT REPAIR]}} \\[2pt]

\textbf{Detected failure type.} \\
\{\texttt{failure\_type}\} \\[3pt]

\textbf{Assigned retrieval labels.} \\
\{\texttt{retrieval\_labels}\} \\[3pt]

\textbf{General repair advice.} \\
\{\texttt{failure\_level\_advice}\} \\
\{\texttt{label\_specific\_advice}\} \\[3pt]

\textbf{Why this was detected.} \\
\{\texttt{local\_evidence\_bullets}\} \\[3pt]

\textbf{Detector signal.} \\
\{\texttt{detector\_diagnosis\_and\_constraint}\} \\[3pt]

\textbf{Insights from previous successful recoveries.} \\
\{\texttt{top\_k\_retrieved\_playbook\_entries\_or\_empty\_message}\} \\[3pt]

\textbf{Continuation instruction.} \\
Continue the original task from the corrected next step. Keep following the
original task protocol. \\[3pt]

\textbf{Required action format.} \\
\{\texttt{schema\_description}\}
\end{tcolorbox}

\begin{tcolorbox}[
  enhanced,
  breakable,
  title={Sentry escape verifier prompt skeleton},
  colback=gray!3,
  colframe=black!55,
  coltitle=white,
  colbacktitle=black!55,
  fonttitle=\bfseries,
  boxrule=0.5pt,
  arc=2pt,
  left=4pt,
  right=4pt,
  top=4pt,
  bottom=4pt
]
\textcolor{black!70}{\texttt{[SENTRY ESCAPE VERIFIER]}} \\[2pt]

\textbf{Role.} \\
You are the Sentry recovery verifier. Your job is to judge whether the
intervention fixed the exact failure that triggered the repair. Judge the
specific retrieval labels and resolution criteria, not generic task success
alone. \\[3pt]

\textbf{Task objective.} \\
\{\texttt{task\_objective}\} \\[3pt]

\textbf{Triggering failure.} \\
\{\texttt{failure\_type}\} \\
\{\texttt{retrieval\_labels}\} \\
\{\texttt{trigger\_evidence}\} \\[3pt]

\textbf{Recovery trajectory.} \\
\{\texttt{post\_intervention\_trajectory}\} \\[3pt]

\textbf{Decision criteria.} \\
A failure is resolved if the agent stops the diagnosed behavior and resumes
valid, task-relevant, observation-grounded progress. A failure is not resolved
if the same failure pattern persists, the agent drifts further, or the recovery
introduces unsupported assumptions. \\[3pt]

\textbf{Output format.} \\
Return resolved or unresolved, a short justification, and the evidence used for
the judgment.
\end{tcolorbox}

\begin{tcolorbox}[
  enhanced,
  breakable,
  title={Sentry playbook update prompt skeleton},
  colback=gray!3,
  colframe=black!55,
  coltitle=white,
  colbacktitle=black!55,
  fonttitle=\bfseries,
  boxrule=0.5pt,
  arc=2pt,
  left=4pt,
  right=4pt,
  top=4pt,
  bottom=4pt
]
\textcolor{black!70}{\texttt{[SENTRY PLAYBOOK UPDATE]}} \\[2pt]

\textbf{Role.} \\
You summarize only verified successful repairs into reusable playbook entries.
Do not store generic reflections or advice that is not supported by the local
trajectory. \\[3pt]

\textbf{Verified repair episode.} \\
\{\texttt{failure\_type}\} \\
\{\texttt{retrieval\_labels}\} \\
\{\texttt{trigger\_pattern}\} \\
\{\texttt{repair\_guidance}\} \\
\{\texttt{post\_repair\_evidence}\} \\[3pt]

\textbf{Output format.} \\
Return a concise playbook entry containing the failure type, subtype, trigger
pattern, and repair principle. The entry should be specific enough to help a
future similar failure, but not so narrow that it only applies to the original
trajectory.
\end{tcolorbox}

\subsubsection{Example playbook}
\label{playbook}

We show an example Sentry playbook from WebShop.

\paragraph{Progress failures.}

\begin{itemize}[leftmargin=1.4em]
\item \texttt{[rep-00001]} \textbf{Repetition or looping.}
If the same search query returns the same irrelevant result page, change one
meaningful query variable from the user request rather than repeating the query.

\item \texttt{[rep-00002]} \textbf{Repetition or looping.}
If clicking an item and returning to search produces no new useful evidence, do
not reopen the same item; inspect a different visible candidate or reformulate
the query.

\item \texttt{[rep-00003]} \textbf{Repetition or looping.}
If navigation cycles between results, item page, and back-to-search without
reward or new attributes, summarize what is known and take a state-changing
action.

\item \texttt{[stall-00004]} \textbf{Planning stall.}
When the current page already exposes candidate items, convert the goal into
one concrete action: inspect a candidate, choose an option, or buy if
requirements are satisfied.

\item \texttt{[stall-00005]} \textbf{Planning stall.}
If the agent says it needs to check size, color, material, or price, the next
action should be the specific click or search that can verify that attribute.

\item \texttt{[stall-00006]} \textbf{Planning stall.}
When no perfect result is obvious, choose the next most diagnostic action
instead of continuing to restate the shopping goal.

\item \texttt{[explore-00007]} \textbf{Over-exploration.}
When a plausible candidate is visible, inspect it before issuing another broad
search.

\item \texttt{[explore-00008]} \textbf{Over-exploration.}
When an item page already shows enough evidence for the requested constraints,
commit instead of opening more unrelated candidates.

\item \texttt{[explore-00009]} \textbf{Over-exploration.}
Near the action budget limit, prefer narrowing or committing over another
generic search.

\item \texttt{[explore-00010]} \textbf{Over-exploration.}
If several visible candidates share the requested category, compare those
candidates directly instead of restarting from the search page.

\item \texttt{[term-00011]} \textbf{Termination miscalibration.}
If \texttt{Buy Now} is visible and the observed item satisfies the hard
constraints, use \texttt{click[Buy Now]} instead of searching or navigating
away.

\item \texttt{[term-00012]} \textbf{Termination miscalibration.}
If the task is not yet purchased, do not stop merely because a good candidate
was found; finish with the valid purchase action.

\item \texttt{[term-00013]} \textbf{Termination miscalibration.}
If only one or two actions remain and a valid commit action is available,
prioritize committing over further exploration.
\end{itemize}

\paragraph{Reasoning \& grounding failures.}

\begin{itemize}[leftmargin=1.4em]
\item \texttt{[hall-00014]} \textbf{Hallucination.}
Do not infer size, color, material, or compatibility from the title alone when
the observation has not shown that attribute.

\item \texttt{[hall-00015]} \textbf{Hallucination.}
If an attribute is missing from the item page, verify it through visible options
or choose another candidate rather than claiming it is satisfied.

\item \texttt{[hall-00016]} \textbf{Hallucination.}
Do not claim ``no matching item exists'' after only checking one irrelevant
result page.

\item \texttt{[drift-00017]} \textbf{Objective drift.}
Keep the original shopping constraints as hard filters; do not buy a nearby
product just because it is cheap or generally similar.

\item \texttt{[drift-00018]} \textbf{Objective drift.}
If the current candidate matches the category but violates size, color, or price
constraints, reject it and continue with the original objective.

\item \texttt{[drift-00019]} \textbf{Objective drift.}
Do not optimize for popularity, broad relevance, or low price when the user
requested specific attributes.

\item \texttt{[mismatch-00020]} \textbf{Reasoning-action mismatch.}
If the reasoning says ``check the item details,'' the action should inspect the
candidate, not go back to search or buy immediately.

\item \texttt{[mismatch-00021]} \textbf{Reasoning-action mismatch.}
If the reasoning identifies one candidate as promising, the click target should
match that candidate.

\item \texttt{[mismatch-00022]} \textbf{Reasoning-action mismatch.}
If the observation shows the required control, the next action should use that
visible control rather than an unrelated search.
\end{itemize}

%% file: figures/detailed_example_figure_appendix.tex
% Compact stacked Figure 4 artwork ONLY: no figure environment, caption or label.
% Replace your existing fig_cases_v3_tikz.tex with this file.
% Keep the figure environment, caption and label in the parent document:
% \begin{figure}[p]
%   \centering
%   \input{fig_cases_v3_tikz.tex}
%   \caption{...}
%   \label{fig:failure_knowledge_cases_detailed}
% \end{figure}
%
% Required preamble packages are the same as for the original drawing:
% \usepackage[T1]{fontenc}
% \usepackage[utf8]{inputenc}
% \usepackage{times,tikz,graphicx,textcomp}
% No additional caption package or ContinuedFloat is required by this file.
%
% Panels (a)/(b) are stacked at full column width. Blank areas are shortened;
% the original text proportions and key task evidence are retained.

\definecolor{c111111}{HTML}{111111}
\definecolor{c1E1E1E}{HTML}{1E1E1E}
\definecolor{c222222}{HTML}{222222}
\definecolor{c2E7D32}{HTML}{2E7D32}
\definecolor{c2F2F2F}{HTML}{2F2F2F}
\definecolor{c333333}{HTML}{333333}
\definecolor{c3B6EA8}{HTML}{3B6EA8}
\definecolor{c4A6FA5}{HTML}{4A6FA5}
\definecolor{c6B6F76}{HTML}{6B6F76}
\definecolor{c6E6E6E}{HTML}{6E6E6E}
\definecolor{c7A4E00}{HTML}{7A4E00}
\definecolor{c7BC47F}{HTML}{7BC47F}
\definecolor{c8A5A00}{HTML}{8A5A00}
\definecolor{c8C93A0}{HTML}{8C93A0}
\definecolor{c9AA0A8}{HTML}{9AA0A8}
\definecolor{cA9C4E8}{HTML}{A9C4E8}
\definecolor{cB3261E}{HTML}{B3261E}
\definecolor{cB7791F}{HTML}{B7791F}
\definecolor{cB9CBE3}{HTML}{B9CBE3}
\definecolor{cC5CCD6}{HTML}{C5CCD6}
\definecolor{cD0D4DA}{HTML}{D0D4DA}
\definecolor{cD3D8DF}{HTML}{D3D8DF}
\definecolor{cD5DEEA}{HTML}{D5DEEA}
\definecolor{cDCE2EA}{HTML}{DCE2EA}
\definecolor{cE24B4A}{HTML}{E24B4A}
\definecolor{cE2E2E2}{HTML}{E2E2E2}
\definecolor{cE4F3E2}{HTML}{E4F3E2}
\definecolor{cE8A33D}{HTML}{E8A33D}
\definecolor{cEAF2FB}{HTML}{EAF2FB}
\definecolor{cEF9A94}{HTML}{EF9A94}
\definecolor{cEFEFEF}{HTML}{EFEFEF}
\definecolor{cF0B54D}{HTML}{F0B54D}
\definecolor{cF1F2F4}{HTML}{F1F2F4}
\definecolor{cF3F4F6}{HTML}{F3F4F6}
\definecolor{cF6F6F6}{HTML}{F6F6F6}
\definecolor{cF7F9FC}{HTML}{F7F9FC}
\definecolor{cFBEFA3}{HTML}{FBEFA3}
\definecolor{cFDE9E7}{HTML}{FDE9E7}
\definecolor{cFFE2A8}{HTML}{FFE2A8}
\definecolor{cFFF1D6}{HTML}{FFF1D6}
\definecolor{cFFF4DE}{HTML}{FFF4DE}
\definecolor{cFFFFFF}{HTML}{FFFFFF}
\providecommand{\fS}[1]{\fontsize{#1pt}{#1pt}\selectfont\sffamily}
\providecommand{\fM}[1]{\fontsize{#1pt}{#1pt}\selectfont\ttfamily}
\newcommand{\SentryCompactACE}{%
  % \resizebox{\linewidth}{!}{%
  \resizebox{0.9\linewidth}{!}{%
    \begin{tikzpicture}[x=1pt,y=-1pt,line cap=butt,line join=miter]
\node[anchor=base west,inner sep=0pt,outer sep=0pt,text=c111111,font=\fS{18}\bfseries] at (12,20) {(a) Context evolution (ACE)};
\path[draw=cC5CCD6,line width=1pt] (12,33) -- (712,33);
\begin{scope}[yshift=53pt]
\path[fill=cF3F4F6, rounded corners=12pt] (12,98) rectangle (304,547);
\node[anchor=base west, inner sep=0pt, outer sep=0pt, text=c6B6F76, font=\fS{12.5}\bfseries] at (18,119) {Task 6594 \textperiodcentered{} earlier};
\path[draw=cD3D8DF, line width=1pt] (135.5,115) -- (298,115);
\path[fill=cF1F2F4, rounded corners=10pt] (60,130) rectangle (298,253);
\node[anchor=base west, inner sep=0pt, outer sep=0pt, text=c1E1E1E, font=\fM{13.5}] at (74,152.8) {search{[}women's pumps{]}};
\path[fill=cFFFFFF, draw=cDCE2EA, line width=1pt, rounded corners=7pt] (72,163) rectangle (286,213);
\node[anchor=base west, inner sep=0pt, outer sep=0pt, text=c222222, font=\fS{14}] at (84,184.5) {Irrelevant results only};
\node[anchor=base west, inner sep=0pt, outer sep=0pt, text=cB3261E, font=\fS{13}\bfseries] at (84,202.5) {query repeated 20 times};
\path[fill=cFDE9E7, draw=cEF9A94, line width=1pt, rounded corners=11pt] (214.6,222) rectangle (286,244);
\node[anchor=base west, inner sep=0pt, outer sep=0pt, text=cB3261E, font=\fS{12.5}\bfseries] at (224.6,237.2) {reward 0};
\path[fill=cFFFFFF, draw=cD5DEEA, line width=1pt] (36,152) circle[radius=17pt];
\path[draw=c4A6FA5, line width=1.4pt] (36,140.5) -- (36,144.5);
\path[fill=cE24B4A] (36,139.5) circle[radius=1.8pt];
\path[fill=cFFFFFF, draw=c4A6FA5, line width=1.4pt, rounded corners=3pt] (28,144.5) rectangle (44,157);
\path[fill=c4A6FA5, rounded corners=1pt] (25,148.5) rectangle (27.6,154);
\path[fill=c4A6FA5, rounded corners=1pt] (44.4,148.5) rectangle (47,154);
\path[fill=c4A6FA5] (32.6,149.8) circle[radius=1.4pt];
\path[fill=c4A6FA5] (39.4,149.8) circle[radius=1.4pt];
\path[draw=c4A6FA5, line width=1.3pt, line cap=round] (32.6,153.6) .. controls (34.87,155.2) and (37.13,155.2) .. (39.4,153.6);
\path[fill=cFFFFFF, draw=cA9C4E8, line width=1.2pt, rounded corners=10pt] (18,262) rectangle (298,466);
\node[anchor=base west, inner sep=0pt, outer sep=0pt, text=c3B6EA8, font=\fS{14}\bfseries] at (32,289) {ACE playbook --- 3 recovery rules};
\node[anchor=base west, inner sep=0pt, outer sep=0pt, text=c3B6EA8, font=\fS{14}\bfseries] at (32,309) {learned from this failure};
\node[anchor=base west, inner sep=0pt, outer sep=0pt, text=c6B6F76, font=\fS{12.5}] at (32,327.2) {injected into the agent's context before the};
\node[anchor=base west, inner sep=0pt, outer sep=0pt, text=c6B6F76, font=\fS{12.5}] at (32,345.2) {next task starts};
\path[fill=c3B6EA8] (38,364) circle[radius=2.2pt];
\path[fill=cFBEFA3, rounded corners=3pt] (46,357) rectangle (144.1,375);
\node[anchor=base west, inner sep=0pt, outer sep=0pt, text=c222222, font=\fS{14}] at (48,369) {no viable result \ensuremath{\rightarrow} broaden the query};
\path[fill=c3B6EA8] (38,384) circle[radius=2.2pt];
\path[fill=cFBEFA3, rounded corners=3pt] (46,377) rectangle (173.7,395);
\node[anchor=base west, inner sep=0pt, outer sep=0pt, text=c222222, font=\fS{14}] at (48,389) {no match on page 1 \ensuremath{\rightarrow} check the next};
\node[anchor=base west, inner sep=0pt, outer sep=0pt, text=c222222, font=\fS{14}] at (48,409) {pages};
\path[fill=c3B6EA8] (38,424) circle[radius=2.2pt];
\path[fill=cFBEFA3, rounded corners=3pt] (46,417) rectangle (176.8,435);
\node[anchor=base west, inner sep=0pt, outer sep=0pt, text=c222222, font=\fS{14}] at (48,429) {results still irrelevant \ensuremath{\rightarrow} refine the};
\node[anchor=base west, inner sep=0pt, outer sep=0pt, text=c222222, font=\fS{14}] at (48,449) {query};
\path[fill=cF7F9FC, rounded corners=12pt] (328,98) rectangle (712,547);
\node[anchor=base west, inner sep=0pt, outer sep=0pt, text=c6B6F76, font=\fS{12.5}\bfseries] at (334,119) {Task 4494 \textperiodcentered{} later};
\path[draw=cD3D8DF, line width=1pt] (440.3,115) -- (706,115);
\path[fill=cEAF2FB, rounded corners=10pt] (334,130) rectangle (656,190);
\node[anchor=base west, inner sep=0pt, outer sep=0pt, text=c222222, font=\fS{15}] at (348,156) {I need high-quality hair extensions, 20 inch,};
\node[anchor=base west, inner sep=0pt, outer sep=0pt, text=c222222, font=\fS{15}] at (348,175) {under \$80.};
\path[fill=cFFFFFF, draw=cE2E2E2, line width=1pt] (688,152) circle[radius=17pt];
\path[fill=c2F2F2F] (688,147.5) circle[radius=4.3pt];
\path[fill=c2F2F2F] (679.5,161) -- (679.57,159.89) -- (679.79,158.8) -- (680.15,157.75) -- (680.64,156.75) -- (681.26,155.83) -- (681.99,154.99) -- (682.83,154.26) -- (683.75,153.64) -- (684.75,153.15) -- (685.8,152.79) -- (686.89,152.57) -- (688,152.5) -- (689.11,152.57) -- (690.2,152.79) -- (691.25,153.15) -- (692.25,153.64) -- (693.17,154.26) -- (694.01,154.99) -- (694.74,155.83) -- (695.36,156.75) -- (695.85,157.75) -- (696.21,158.8) -- (696.43,159.89) -- (696.5,161) -- cycle;
\path[fill=cF1F2F4, rounded corners=10pt] (376,199) rectangle (706,351);
\node[anchor=base west, inner sep=0pt, outer sep=0pt, text=c1E1E1E, font=\fM{13.5}] at (390,221.8) {search{[}hair extensions 20 inch{]}};
\path[fill=cFFFFFF, draw=cDCE2EA, line width=1pt, rounded corners=7pt] (388,232) rectangle (694,337);
\node[anchor=base west, inner sep=0pt, outer sep=0pt, text=c6B6F76, font=\fS{11.5}] at (400,250.2) {WebShop \textperiodcentered{} search results};
\path[fill=cFBEFA3, rounded corners=3pt] (398,258.5) rectangle (658.5,276.5);
\node[anchor=base west, inner sep=0pt, outer sep=0pt, text=c222222, font=\fS{14}] at (400,270.5) {1. QAZPL Deep Curly Virgin Human Hair,};
\path[fill=cFBEFA3, rounded corners=3pt] (398,277.5) rectangle (579.4,295.5);
\node[anchor=base west, inner sep=0pt, outer sep=0pt, text=c222222, font=\fS{14}] at (400,289.5) {Grade 8A, 20 inch --- \$79.63};
\path[draw=c2E7D32, line width=1.8pt, line cap=round, line join=round] (400,303) -- (403.5,306.5) -- (409.5,299.5);
\node[anchor=base west, inner sep=0pt, outer sep=0pt, text=c2E7D32, font=\fS{13}\bfseries] at (415,307.5) {20 inch \textperiodcentered{} under \$80 \textperiodcentered{} high quality};
\node[anchor=base west, inner sep=0pt, outer sep=0pt, text=c9AA0A8, font=\fS{14}] at (400,326.5) {2. \ldots{}    3. \ldots{}    4. \ldots{}};
\path[fill=cFFFFFF, draw=cD5DEEA, line width=1pt] (352,221) circle[radius=17pt];
\path[draw=c4A6FA5, line width=1.4pt] (352,209.5) -- (352,213.5);
\path[fill=cE24B4A] (352,208.5) circle[radius=1.8pt];
\path[fill=cFFFFFF, draw=c4A6FA5, line width=1.4pt, rounded corners=3pt] (344,213.5) rectangle (360,226);
\path[fill=c4A6FA5, rounded corners=1pt] (341,217.5) rectangle (343.6,223);
\path[fill=c4A6FA5, rounded corners=1pt] (360.4,217.5) rectangle (363,223);
\path[fill=c4A6FA5] (348.6,218.8) circle[radius=1.4pt];
\path[fill=c4A6FA5] (355.4,218.8) circle[radius=1.4pt];
\path[draw=c4A6FA5, line width=1.3pt, line cap=round] (348.6,222.6) .. controls (350.87,224.2) and (353.13,224.2) .. (355.4,222.6);
\path[fill=cF1F2F4, rounded corners=10pt] (376,360) rectangle (706,490);
\path[fill=cFBEFA3, rounded corners=3pt] (388,371.5) rectangle (622.7,390.5);
\node[anchor=base west, inner sep=0pt, outer sep=0pt, text=c222222, font=\fS{15}] at (390,384.5) {Item 1 fits the description perfectly. Let me};
\node[anchor=base west, inner sep=0pt, outer sep=0pt, text=c222222, font=\fS{15}] at (390,404.5) {also go through item 2, item 3, item 4};
\node[anchor=base west, inner sep=0pt, outer sep=0pt, text=c222222, text opacity=0.4, font=\fS{15}] at (390,424.5) {to make sure there is nothing better\ldots{}};
\path[fill=cB3261E] (645,412) rectangle (647.5,427);
\path[fill=cFDE9E7, draw=cEF9A94, line width=1pt, rounded corners=11pt]
    (549,433) rectangle (694,455);

\node[anchor=base west, inner sep=0pt, outer sep=0pt,
      text=cB3261E, font=\fS{12.5}\bfseries]
    at (559,448.2) {no action, reward = 0};
\path[fill=cFFFFFF, draw=cD5DEEA, line width=1pt] (352,382) circle[radius=17pt];
\path[draw=c4A6FA5, line width=1.4pt] (352,370.5) -- (352,374.5);
\path[fill=cE24B4A] (352,369.5) circle[radius=1.8pt];
\path[fill=cFFFFFF, draw=c4A6FA5, line width=1.4pt, rounded corners=3pt] (344,374.5) rectangle (360,387);
\path[fill=c4A6FA5, rounded corners=1pt] (341,378.5) rectangle (343.6,384);
\path[fill=c4A6FA5, rounded corners=1pt] (360.4,378.5) rectangle (363,384);
\path[fill=c4A6FA5] (348.6,379.8) circle[radius=1.4pt];
\path[fill=c4A6FA5] (355.4,379.8) circle[radius=1.4pt];
\path[draw=c4A6FA5, line width=1.3pt, line cap=round] (348.6,383.6) .. controls (350.87,385.2) and (353.13,385.2) .. (355.4,383.6);
\path[fill=cFFFFFF, draw=c9AA0A8, line width=1.2pt, dash pattern=on 5pt off 4pt, rounded corners=10pt] (376,499) rectangle (541.6,535);
\node[anchor=base west, inner sep=0pt, outer sep=0pt, text=c6B6F76, font=\fM{13.5}] at (390,521.5) {click{[}B08X4YD5WH{]}};
\node[anchor=base west, inner sep=0pt, outer sep=0pt, text=c6B6F76, font=\fS{12.5}\itshape] at (553.6,521) {correct action --- never taken};
\path[fill=cFFFFFF, draw=cD5DEEA, line width=1pt, opacity=0.45] (352,517) circle[radius=17pt];
\path[draw=c4A6FA5, line width=1.4pt, opacity=0.45] (352,505.5) -- (352,509.5);
\path[fill=cE24B4A, opacity=0.45] (352,504.5) circle[radius=1.8pt];
\path[fill=cFFFFFF, draw=c4A6FA5, line width=1.4pt, opacity=0.45, rounded corners=3pt] (344,509.5) rectangle (360,522);
\path[fill=c4A6FA5, opacity=0.45, rounded corners=1pt] (341,513.5) rectangle (343.6,519);
\path[fill=c4A6FA5, opacity=0.45, rounded corners=1pt] (360.4,513.5) rectangle (363,519);
\path[fill=c4A6FA5, opacity=0.45] (348.6,514.8) circle[radius=1.4pt];
\path[fill=c4A6FA5, opacity=0.45] (355.4,514.8) circle[radius=1.4pt];
\path[draw=c4A6FA5, line width=1.3pt, line cap=round, opacity=0.45] (348.6,518.6) .. controls (350.87,520.2) and (353.13,520.2) .. (355.4,518.6);

\path[draw=c3B6EA8,line width=2pt] (305,363.5) -- (321,363.5);
\path[draw=c3B6EA8,line width=2pt,line cap=round,line join=round] (320,358.5) -- (328,363.5) -- (320,368.5);
\end{scope}

      \pgfresetboundingbox
      \useasboundingbox (0,0) rectangle (724,500);
    \end{tikzpicture}%
  }%
}

\newcommand{\SentryCompactAgentGuard}{%
  % \resizebox{\linewidth}{!}{%
  \resizebox{0.9\linewidth}{!}{%
    \begin{tikzpicture}[x=1pt,y=-1pt,line cap=butt,line join=miter]
% Compact AgentGuard panel; original x coordinates, y extent 0--500.
% Text retains its original aspect ratio. Colors and font helpers are shared.
\node[anchor=base west, inner sep=0pt, outer sep=0pt, text=c111111, font=\fS{18}\bfseries] at (752,24) {(b) Runtime intervention (AgentGuard)};
\node[anchor=base east, inner sep=0pt, outer sep=0pt, text=c7A4E00, font=\fS{13}] at (1452,23) {external monitor; no memory of past rescues};
\path[fill=cFFF4DE, draw=cF0B54D, line width=1pt, rounded corners=5pt] (752,32) rectangle (1452,55);
\node[anchor=base west, inner sep=0pt, outer sep=0pt, text=c7A4E00, font=\fS{13.5}] at (765,48) {Fixed rule: page unchanged for n steps \ensuremath{\rightarrow} warn ``no progress''};

% Earlier task: a helpful warning, followed by a successful purchase.
\path[fill=cF7F9FC, rounded corners=12pt] (752,63) rectangle (1096,500);
\node[anchor=base west, inner sep=0pt, outer sep=0pt, text=c6B6F76, font=\fS{12.5}\bfseries] at (758,83) {Task 3980 \textperiodcentered{} earlier};
\node[anchor=base west, inner sep=0pt, outer sep=0pt, text=c6B6F76, font=\fS{12.5}] at (758,103) {Repeated identical searches; the agent is stuck.};
\path[fill=cFFF4DE, draw=cF0B54D, line width=1.2pt, rounded corners=10pt] (800,119) rectangle (1090,201);
\node[anchor=base west, inner sep=0pt, outer sep=0pt, text=c222222, font=\fS{15}] at (814,141.5) {No progress detected. Take a};
\node[anchor=base west, inner sep=0pt, outer sep=0pt, text=c222222, font=\fS{15}] at (814,161.5) {state-changing action.};
\path[fill=cE4F3E2, draw=c7BC47F, line width=1pt, rounded corners=11pt] (877.4,170) rectangle (1078,192);
\node[anchor=base west, inner sep=0pt, outer sep=0pt, text=c2E7D32, font=\fS{12.5}\bfseries] at (887.4,185.2) {correct --- the agent was stuck};
\path[fill=cFFF1D6, draw=cE8A33D, line width=1pt] (776,141) circle[radius=17pt];
\path[fill=cFFE2A8, draw=cB7791F, line width=1.3pt, line join=round] (776,132) -- (784.5,147) -- (767.5,147) -- cycle;
\path[draw=c8A5A00, line width=1.6pt, line cap=round] (776,137.5) -- (776,142.5);
\path[fill=c8A5A00] (776,145) circle[radius=0.95pt];

\path[fill=cF1F2F4, rounded corners=10pt] (800,210) rectangle (1090,387);
\node[anchor=base west, inner sep=0pt, outer sep=0pt, text=c1E1E1E, font=\fM{13.5}] at (814,232.8) {click{[}B00O30JLDK{]}};
\node[anchor=base west, inner sep=0pt, outer sep=0pt, text=c1E1E1E, font=\fM{13.5}] at (814,250.8) {click{[}mossy oak country{]}};
\node[anchor=base west, inner sep=0pt, outer sep=0pt, text=c1E1E1E, font=\fM{13.5}] at (814,268.8) {click{[}xx-large big tall{]}};
\path[fill=cFFFFFF, draw=cDCE2EA, line width=1pt, rounded corners=7pt] (812,279) rectangle (1078,329);
\path[fill=cFBEFA3, rounded corners=3pt] (909.1,288.5) rectangle (982.3,306.5);
\node[anchor=base west, inner sep=0pt, outer sep=0pt, text=c222222, font=\fS{14}] at (824,300.5) {Product page unchanged};
\path[draw=c2E7D32, line width=1.8pt, line cap=round, line join=round] (824,314) -- (827.5,317.5) -- (833.5,310.5);
\node[anchor=base west, inner sep=0pt, outer sep=0pt, text=c2E7D32, font=\fS{13}\bfseries] at (839,318.5) {both options registered};
\node[anchor=base west, inner sep=0pt, outer sep=0pt, text=c1E1E1E, font=\fM{13.5}] at (814,347.8) {click{[}Buy Now{]}};
\path[fill=cE4F3E2, draw=c7BC47F, line width=1pt, rounded corners=11pt] (996.2,356) rectangle (1078,378);
\node[anchor=base west, inner sep=0pt, outer sep=0pt, text=c2E7D32, font=\fS{12.5}\bfseries] at (1006.2,371.2) {reward 1.0};
\path[fill=cFFFFFF, draw=cD5DEEA, line width=1pt] (776,232) circle[radius=17pt];
\path[draw=c4A6FA5, line width=1.4pt] (776,220.5) -- (776,224.5);
\path[fill=cE24B4A] (776,219.5) circle[radius=1.8pt];
\path[fill=cFFFFFF, draw=c4A6FA5, line width=1.4pt, rounded corners=3pt] (768,224.5) rectangle (784,237);
\path[fill=c4A6FA5, rounded corners=1pt] (765,228.5) rectangle (767.6,234);
\path[fill=c4A6FA5, rounded corners=1pt] (784.4,228.5) rectangle (787,234);
\path[fill=c4A6FA5] (772.6,229.8) circle[radius=1.4pt];
\path[fill=c4A6FA5] (779.4,229.8) circle[radius=1.4pt];
\path[draw=c4A6FA5, line width=1.3pt, line cap=round] (772.6,233.6) .. controls (774.87,235.2) and (777.13,235.2) .. (779.4,233.6);

\path[fill=cFFFFFF, draw=cEF9A94, line width=1pt, dash pattern=on 4pt off 3pt, rounded corners=10pt] (758,396) rectangle (1090,454);
\node[anchor=base west, inner sep=0pt, outer sep=0pt, text=cB3261E, font=\fS{13}] at (771,413) {Potential lesson: an unchanged page can still};
\node[anchor=base west, inner sep=0pt, outer sep=0pt, text=cB3261E, font=\fS{13}] at (771,430) {reflect progress. AgentGuard stores no lesson};
\node[anchor=base west, inner sep=0pt, outer sep=0pt, text=cB3261E, font=\fS{13}] at (771,447) {from this rescue.};
\node[anchor=base east, inner sep=0pt, outer sep=0pt, text=c8C93A0, font=\fS{11.5}] at (1090,494) {task agent context \textperiodcentered{} task 3980};

% Later task: identical visible page change criterion, but registered options.
\path[fill=cF7F9FC, rounded corners=12pt] (1108,63) rectangle (1452,500);
\node[anchor=base west, inner sep=0pt, outer sep=0pt, text=c6B6F76, font=\fS{12.5}\bfseries] at (1114,83) {Task 4585 \textperiodcentered{} later};
\node[anchor=base west, inner sep=0pt, outer sep=0pt, text=c6B6F76, font=\fS{12.5}\itshape] at (1114,99) {After seven identical searches, the agent};
\node[anchor=base west, inner sep=0pt, outer sep=0pt, text=c6B6F76, font=\fS{12.5}\itshape] at (1114,114) {reopens the same product:};
\path[fill=cF1F2F4, rounded corners=10pt] (1156,119) rectangle (1446,252);
\node[anchor=base west, inner sep=0pt, outer sep=0pt, text=c1E1E1E, font=\fM{13.5}] at (1170,141.8) {click{[}B00O30JLDK{]}};
\node[anchor=base west, inner sep=0pt, outer sep=0pt, text=c1E1E1E, font=\fM{13.5}] at (1170,159.8) {click{[}mossy oak country{]}};
\node[anchor=base west, inner sep=0pt, outer sep=0pt, text=c1E1E1E, font=\fM{13.5}] at (1170,177.8) {click{[}large{]}};
\path[fill=cFFFFFF, draw=cDCE2EA, line width=1pt, rounded corners=7pt] (1168,188) rectangle (1434,238);
\path[fill=cFBEFA3, rounded corners=3pt] (1265.1,197.5) rectangle (1338.3,215.5);
\node[anchor=base west, inner sep=0pt, outer sep=0pt, text=c222222, font=\fS{14}] at (1180,209.5) {Product page unchanged};
\path[draw=c2E7D32, line width=1.8pt, line cap=round, line join=round] (1180,223) -- (1183.5,226.5) -- (1189.5,219.5);
\node[anchor=base west, inner sep=0pt, outer sep=0pt, text=c2E7D32, font=\fS{13}\bfseries] at (1195,227.5) {both options registered};
\path[fill=cFFFFFF, draw=cD5DEEA, line width=1pt] (1132,141) circle[radius=17pt];
\path[draw=c4A6FA5, line width=1.4pt] (1132,129.5) -- (1132,133.5);
\path[fill=cE24B4A] (1132,128.5) circle[radius=1.8pt];
\path[fill=cFFFFFF, draw=c4A6FA5, line width=1.4pt, rounded corners=3pt] (1124,133.5) rectangle (1140,146);
\path[fill=c4A6FA5, rounded corners=1pt] (1121,137.5) rectangle (1123.6,143);
\path[fill=c4A6FA5, rounded corners=1pt] (1140.4,137.5) rectangle (1143,143);
\path[fill=c4A6FA5] (1128.6,138.8) circle[radius=1.4pt];
\path[fill=c4A6FA5] (1135.4,138.8) circle[radius=1.4pt];
\path[draw=c4A6FA5, line width=1.3pt, line cap=round] (1128.6,142.6) .. controls (1130.87,144.2) and (1133.13,144.2) .. (1135.4,142.6);

\path[fill=cFFF4DE, draw=cF0B54D, line width=1.2pt, rounded corners=10pt] (1156,261) rectangle (1446,345);
\node[anchor=base west, inner sep=0pt, outer sep=0pt, text=c222222, font=\fS{15}] at (1170,283.5) {No progress detected. Take a};
\node[anchor=base west, inner sep=0pt, outer sep=0pt, text=c222222, font=\fS{15}] at (1170,303.5) {state-changing action.};
\path[fill=cFDE9E7, draw=cEF9A94, line width=1pt, rounded corners=7pt] (1165,309) rectangle (1434,341);
\node[anchor=base west, inner sep=0pt, outer sep=0pt, text=cB3261E, font=\fS{12.5}\bfseries] at (1172,322) {wrong --- options are registered;};
\node[anchor=base west, inner sep=0pt, outer sep=0pt, text=cB3261E, font=\fS{12.5}\bfseries] at (1172,336) {the correct action is to buy.};
\path[fill=cFFF1D6, draw=cE8A33D, line width=1pt] (1132,283) circle[radius=17pt];
\path[fill=cFFE2A8, draw=cB7791F, line width=1.3pt, line join=round] (1132,274) -- (1140.5,289) -- (1123.5,289) -- cycle;
\path[draw=c8A5A00, line width=1.6pt, line cap=round] (1132,279.5) -- (1132,284.5);
\path[fill=c8A5A00] (1132,287) circle[radius=0.95pt];

\path[fill=cF1F2F4, rounded corners=10pt] (1156,354) rectangle (1446,486);
\node[anchor=base west, inner sep=0pt, outer sep=0pt, text=c1E1E1E, font=\fM{13.5}] at (1170,376.8) {click{[}\textless{} Prev{]}};
\node[anchor=base west, inner sep=0pt, outer sep=0pt, text=c1E1E1E, font=\fM{13.5}] at (1170,394.8) {click{[}B00O30JLDK{]}};
\node[anchor=base west, inner sep=0pt, outer sep=0pt, text=c1E1E1E, font=\fM{13.5}] at (1170,412.8) {click{[}Buy Now{]}};
\path[fill=cFFFFFF, draw=cDCE2EA, line width=1pt, rounded corners=7pt] (1168,423) rectangle (1434,451);
\path[fill=cFBEFA3, rounded corners=3pt] (1277.6,432.5) rectangle (1355.9,450.5);
\node[anchor=base west, inner sep=0pt, outer sep=0pt, text=c222222, font=\fS{14}] at (1180,444.5) {Purchased with options = \{ \}};
\path[fill=cFDE9E7, draw=cEF9A94, line width=1pt, rounded corners=11pt] (1352.2,460) rectangle (1434,482);
\node[anchor=base west, inner sep=0pt, outer sep=0pt, text=cB3261E, font=\fS{12.5}\bfseries] at (1362.2,475.2) {reward 0.6};
\path[fill=cFFFFFF, draw=cD5DEEA, line width=1pt] (1132,376) circle[radius=17pt];
\path[draw=c4A6FA5, line width=1.4pt] (1132,364.5) -- (1132,368.5);
\path[fill=cE24B4A] (1132,363.5) circle[radius=1.8pt];
\path[fill=cFFFFFF, draw=c4A6FA5, line width=1.4pt, rounded corners=3pt] (1124,368.5) rectangle (1140,381);
\path[fill=c4A6FA5, rounded corners=1pt] (1121,372.5) rectangle (1123.6,378);
\path[fill=c4A6FA5, rounded corners=1pt] (1140.4,372.5) rectangle (1143,378);
\path[fill=c4A6FA5] (1128.6,373.8) circle[radius=1.4pt];
\path[fill=c4A6FA5] (1135.4,373.8) circle[radius=1.4pt];
\path[draw=c4A6FA5, line width=1.3pt, line cap=round] (1128.6,377.6) .. controls (1130.87,379.2) and (1133.13,379.2) .. (1135.4,377.6);
\node[anchor=base east, inner sep=0pt, outer sep=0pt, text=c8C93A0, font=\fS{11.5}] at (1446,494) {task agent context \textperiodcentered{} task 4585};

      \pgfresetboundingbox
      \useasboundingbox (740,0) rectangle (1464,500);
    \end{tikzpicture}%
  }%
}

\begingroup
\setlength{\parskip}{0pt}
\SentryCompactACE\par
\vspace{4pt}
\SentryCompactAgentGuard\par
\endgroup